\documentclass[10pt,twocolumn,letterpaper]{article}

\usepackage[pagenumbers]{cvpr} 

\usepackage[dvipsnames]{xcolor}

\definecolor{cvprblue}{rgb}{0.21,0.49,0.74}
\usepackage{multirow}
\usepackage{bbm}
\usepackage{array}
\usepackage{colortbl}
\newcolumntype{P}[1]{>{\centering\arraybackslash}p{#1}}
\usepackage[pagebackref,breaklinks,colorlinks,citecolor=cvprblue]{hyperref}

\def\paperID{6960} 
\def\confName{CVPR}
\def\confYear{2024}

\title{Style Blind Domain Generalized Semantic Segmentation \\ via Covariance Alignment and Semantic Consistence Contrastive Learning}

\author{Woo-Jin Ahn$^{1}$\quad Geun-Yeong Yang$^{1}$\quad Hyun-Duck Choi$^{2}$\thanks{Corresponding Authors} \quad Myo-Taeg Lim$^{1}$\footnotemark[1] \\
$^{1}$Korea University \quad
$^{2}$Chonnam National University\\
{\tt\small $^{1}$\{wjahn,hggofficial,mlim\}@korea.ac.kr \; $^{2}$ducky.choi@jnu.ac.kr}}

\begin{document}
\maketitle
\begin{abstract}
Deep learning models for semantic segmentation often experience performance degradation when deployed to unseen target domains unidentified during the training phase. This is mainly due to variations in image texture (\ie style) from different data sources. To tackle this challenge, existing domain generalized semantic segmentation (DGSS) methods attempt to remove style variations from the feature. However, these approaches struggle with the entanglement of style and content, which may lead to the unintentional removal of crucial content information, causing performance degradation. This study addresses this limitation by proposing BlindNet, a novel DGSS approach that blinds the style without external modules or datasets. The main idea behind our proposed approach is to alleviate the effect of style in the encoder whilst facilitating robust segmentation in the decoder. To achieve this, BlindNet comprises two key components: covariance alignment and semantic consistency contrastive learning. Specifically, the covariance alignment trains the encoder to uniformly recognize various styles and preserve the content information of the feature, rather than removing the style-sensitive factor. Meanwhile, semantic consistency contrastive learning enables the decoder to construct discriminative class embedding space and disentangles features that are vulnerable to misclassification. Through extensive experiments, our approach outperforms existing DGSS methods, exhibiting robustness and superior performance for semantic segmentation on unseen target domains. The code is available at \url{https://github.com/root0yang/BlindNet}.
\end{abstract}
    
\section{Introduction}
\label{sec:intro}

\begin{figure}[t]
  \centering
  \includegraphics[width=\linewidth]{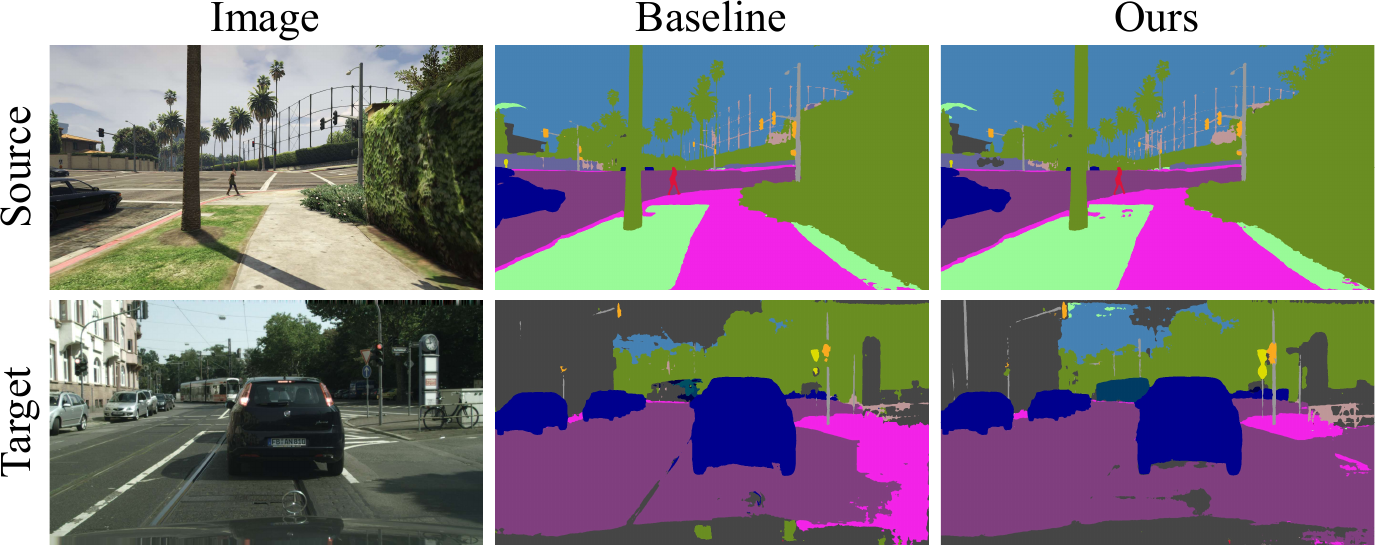}
  \caption{Comparison of semantic segmentation results between the baseline (DeepLabV3+ with ResNet50 backbone) and our BlindNet. Both models are trained on the source domain (GTAV~\cite{richter2016playing}) and tested on the target domain (Cityscapes~\cite{cordts2016cityscapes}).}
  \label{fig:intro}
\end{figure}

Semantic segmentation, a technique that classifies each pixel in an image into predefined categories, has garnered significant attention due to its potential applications in various fields. Particularly,  it plays a crucial role in autonomous driving~\cite{hu2023planning,bartoccioni2023lara} and robotic systems~\cite{onozuka2021autonomous,nilsson2021embodied}. Besides, with the advent of large datasets, deep neural networks have emerged as a trending approach for semantic segmentation tasks, achieving impressive results~\cite{xie2021segformer,cheng2022masked,ronneberger2015u,chen2017rethinking}. However, there remains a major bottleneck detailing the meticulous and labor-intensive process of dataset labeling. More particularly, this process not only consumes time but also poses economic challenges~\cite{cordts2016cityscapes, sakaridis2021acdc}. To address this challenge, synthetic datasets have emerged as a compelling alternative. Specifically, these datasets, generated using three-dimensional (3D) rendering techniques, offer vast amounts of easily accessible data, eliminating the need for manual labeling~\cite{richter2016playing, ros2016synthia}. However, a challenge arises when models trained on synthetic datasets are deployed in real-world scenarios. More precisely, a  domain shift problem arises due to style factor discrepancies (\eg texture, illumination, and image quality) between synthetic and real-world data, which affects the performance of the model, as shown in Fig.~\ref{fig:intro}.

To address the domain shift problem, domain adaptation semantic segmentation (DASS) has been introduced~\cite{li2020content,hoyer2023improving,hoffman2018cycada,vu2019advent,zou2018unsupervised,huang2018multimodal}. Specifically, DASS  aims to bridge the gap between source and target domains by aligning their data distributions. However, a significant limitation of DASS is its dependency on the target domain during training. For DASS to function effectively, target domain samples must be available during the training phase.

Meanwhile,  domain generalized semantic segmentation (DGSS) has been introduced as an alternative approach to tackle the domain shift problem. Unlike DASS, DGSS is trained only with the source domain, aiming to extract domain-invariant features. To achieve this, two main techniques have been employed: domain randomization (DR) and feature normalization (FN).

DR augments the training set by introducing variability, either by altering the image style~\cite{peng2021global,yue2019domain} or by modifying the feature representation~\cite{lee2022wildnet,wu2022siamdoge}. By exposing the model to a wider variety of styles via DR, the network is less likely to overfit to the specific styles present in the training data. Consequently, the robustness of the model is improved, making it more adept at generalizing to new, unseen domains. Nevertheless, a crucial limitation of DR is its significant dependence on auxiliary domains.



FN methods, converse to DR, regularize the features to prevent the model from overfitting to the distinct styles or characteristics of the training data. This is achieved by removing domain-specific style information using feature statistics, such as instance normalization~\cite{pan2018two} or whitening transformation~\cite{choi2021robustnet,peng2022semantic,pan2019switchable,huang2021fsdr}. While these approaches effectively remove style-related information, they simultaneously pose the challenge of removing semantic content because content and style information are entangled. Consequently, the model fails to capture the essential patterns or features required for accurate segmentation prediction.

To address this problem, we propose BlindNet, a model that blinds the style within the encoder and improves the robustness of the decoder, without requiring auxiliary datasets or external modules. Specifically, our proposed BlindNet consists of two components: covariance alignment for the encoder and semantic consistency contrastive learning for the decoder. Precisely, the covariance alignment facilitates the generation of style-invariant features with the proposed covariance matching loss function (CML) and the cross-covariance loss function (CCL). 
Specifically, CML mitigates the effects of style variations, while CCL focuses on preserving content information, effectively addressing the prevalent content information loss observed in the FN method. To further improve the generalization ability, we develop semantic consistency contrastive learning, which consists of class-wise contrastive learning (CWCL) and semantic disentanglement contrastive learning (SDCL). Particularly, the  CWCL constructs a discriminative class embedding space, while SDCL disentangles features of similar classes that often lead to prediction errors. Extensive experiments across various datasets demonstrate that the proposed BlindNet outperforms existing DGSS methods.

Our contributions are summarized as follows:
\begin{itemize}
    \item We propose a covariance alignment within the encoder, comprising  CML and CCL. Specifically, the  CML aims to mitigate the effects of style variations, while CCL ensures the preservation of content information, together facilitating the generation of style-agnostic features.

    \item We propose semantic consistency contrastive learning within the decoder that comprises CWCL and SDCL, utilizing segmentation masks. Specifically,  CWCL generates discriminative embeddings, whereas SDCL disentangles features of similar classes, enhancing the robustness of the model.

    \item Through extensive experiments, we demonstrate the superiority of our approach in DGSS, without the need to alter the network architecture or rely on external datasets.
\end{itemize}

\section{Related Works}

\noindent\textbf{Domain adaptation and generalization for semantic segmentation.} Domain adaptation (DA)  aims at minimizing the distribution discrepancy between different domains, enabling a model to generalize from a source to a target domain. For DASS, adversarial training and cross-domain self-training strategies are commonly used. Particularly, adversarial-based  methods~\cite{hoffman2018cycada,li2019bidirectional} employ generative adversarial networks ~\cite{goodfellow2014generative} to close the feature distribution gap between source and target domains. Meanwhile, cross-domain self-training methods~\cite{zou2018unsupervised,yu2021dast,pan2020unsupervised,hoyer2022daformer} generate pseudo-labels for target domain data using pre-trained models,  and employ them as training data,  thereby expanding the training data and reducing distribution differences between the source and target domain.

Domain generalization (DG) methods ~\cite{li2018deep, volpi2018generalizing, zhou2020domain, li2017learning,li2021progressive} aim to improve the generalization ability of the model without accessing the target domain during training. Since the difference in style of the image is the main cause of the disparity, most existing domain generalized semantic segmentation methods utilize the style information of the image for domain-invariant learning. Feature statistics (\eg mean, variance, covariance, gram matrix, \etc) which are commonly used in style transfer~\cite{gatys2016image,yoo2019photorealistic,li2017universal,huang2017arbitrary} are employed to capture the style information. Interestingly, existing  DGSS methods can be separated into two parts: i) domain randomization to expand the distribution of style or ii) feature normalization to remove style.

DR involves randomizing either the image or its features through stylization to learn domain-invariant features from various styles. For example,  Peng \etal~\cite{peng2022semantic} extended the source domain data by stylizing images in the style of unreal paintings. Similarly, Yue \etal~\cite{yue2019domain} and Huang \etal~\cite{huang2018multimodal} attempted to enhance generalization by synthesizing images with diverse styles in the image space. In another study, Lee  \etal~\cite{lee2022wildnet} adopted ImageNet data~\cite{deng2009imagenet} as wild data and performed randomization via synthesis in the feature space. Meanwhile, Wu  \etal~\cite{wu2022siamdoge} diversified the trainable feature space by mixing the statistics of the feature and its color-jittered feature with Ada-IN~\cite{huang2017arbitrary}.

FN methods aim to remove domain-specific styles from features, extracting only domain-invariant content. For instance,  Pan \etal~\cite{pan2019switchable} first attempted the DGSS method, combining batch normalization (BN)~\cite{ioffe2015batch} and instance normalization  (IN)~\cite{ulyanov2016instance} in the network layer. While BN preserves the content information within discriminative features, IN focuses on removing domain-specific style information from features. In a study, Choi  \etal~\cite{choi2021robustnet} addressed the limitations of previous whitening transformation~\cite{luo2017learning,cho2019image} that can eliminate the content information from the feature. Specifically, they proposed an instance-selective whitening approach designed to remove covariance components that are sensitive to domain shifts. Peng \etal~\cite{peng2021global} developed semantic-aware normalization that performs on class-wise and semantic-aware whitening that aligns channels based on the prediction through group whitening transformation~\cite{cho2019image}. In a study, Xu \etal ~\cite{xu2022dirl} introduced the prior guided attention module and guided feature whitening to re-calibrate the feature and remove domain-specific style effects, respectively. Unlike the FN methods that directly remove the style component, our work explores a covariance alignment method that mitigates the effect of the style’s effect while preserving the content information. Our method achieves the DGSS without any additional modules or auxiliary datasets.

\noindent\textbf{Contrastive Learning.}
Contrastive learning aims to learn representations by maximizing the similarity between positive pairs of samples while minimizing the similarity between negative pairs. In recent years, it has attracted significant attention for its effectiveness in learning discriminative representations across various tasks~\cite{chen2020simple,grill2020bootstrap,he2020momentum,caron2020unsupervised}. Oord \etal~\cite{oord2018representation} were the first to introduce the infoNCE loss, a type of contrastive loss function designed for self-contrastive learning. In a work,  Park \etal~\cite{park2020contrastive} introduced patch-level contrastive learning for image translation, using co-located patches as positive pairs and spatially distant patches as negatives to maintain image context. For the semantic segmentation task, Wang \etal~\cite{wang2021exploring} introduced class-wise contrastive learning to aid the model in learning the embedding space of each class. Specifically, they sampled the classes existing in the images and applied contrastive learning based on the class label. For DGSS, Lee \etal~\cite{lee2022wildnet} adopted contrastive learning to learn the ImageNet information in their model.   Specifically, they set the ImageNet data as wild and applied the contrastive learning method by setting the wild-stylized feature and its closest wild content as positive samples. In another study, Yang \etal~\cite{yang2023generalized} developed multi-level contrastive learning, which designed instance prototypes and class prototypes for contrastive learning. Specifically, they sample each class’s pixel features and apply contrastive learning with a transition-probability ability matrix. Unlike recent DGSS works that embed the original image, we define contrastive learning for the augmented image. Specifically, the proposed method builds a robust embedding space by preserving the semantic consistency of the feature representation across various domains.

\begin{figure*}[h!]
  \centering
  \includegraphics[width=\textwidth]{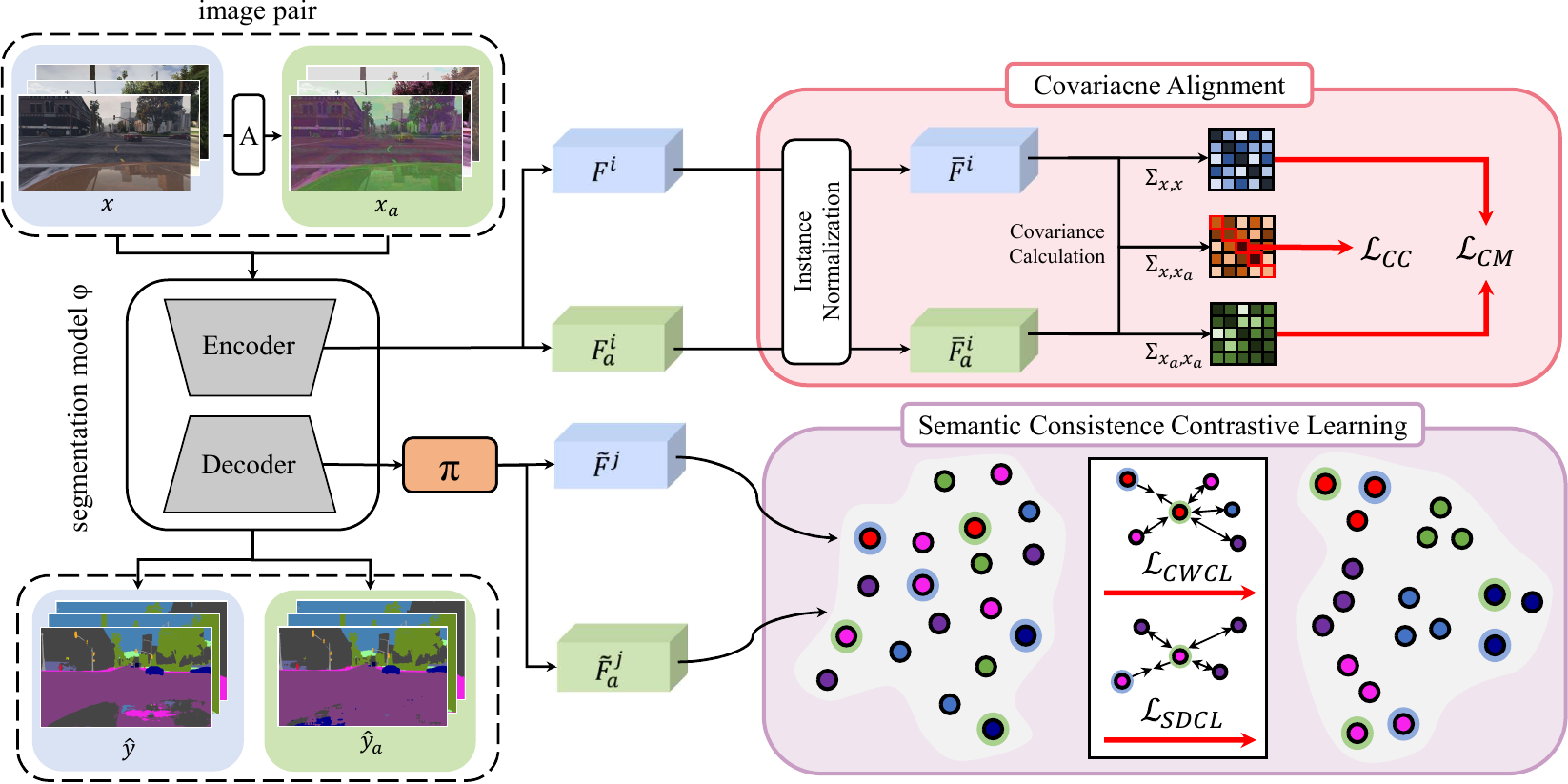}
  \caption{Overview of the proposed BlindNet. The network processes a pair of images - the original image $x$ and its augmented counterpart $x_a$. It employs covariance alignment to treat encoder features and utilizes semantic consistency contrastive learning for the processing of decoder features.}
  \label{fig:overview}
\end{figure*}

\section{Method}
\label{sec:method}

The goal of the proposed method is to train a segmentation model $\varphi$ on a given source domain $S$ and generalize well to the unseen target domain $T$. Precisely, the source domain $S = \{(x,y)\}$ contains a paired image $x \in \mathbb{R}^{H \times W \times 3}$ and segmentation label $y \in \mathbb{R}^{H \times W \times C}$, where $H$, $W$, and $C$ denote the height of the image, the width of the image, and the class number of the segmentation map, respectively. The model $\varphi$ takes an original $x$ and its augmented counterpart $x_a$, which have the same content but different styles, and uses the feature information to enhance its generalization ability.

As shown in Fig.~\ref{fig:overview}, our method leverages the feature information through two main approaches: covariance alignment and semantic consistency contrastive learning. Specifically, the covariance matching ensures that features, having different styles but the same content, contain similar information. Additionally, the semantic consistency contrastive learning embeds the generalized features into discriminative representation based on the segmentation label.

\subsection{Covariance Alignment}
The domain shift in semantic segmentation results from changes in the visual characteristics of the image, known as style. This style information is typically detected in the shallow layers of networks~\cite{pan2018two}, which are the encoders of the segmentation model. Based on this understanding, our method targets the encoder to handle style variations by employing the proposed covariance matching loss and cross-covariance loss function.

\subsubsection*{Covariance Matching Loss}
To train the network to uniformly recognize various styles without removing content information, we introduce the CML. Specifically, the loss aims to minimize the difference between covariance matrices derived from different styles of image features.
Given an image pair $(x,x_a)$, the features from $i^{th}$ block of the encoder are represented as $F^i \in \mathbb{R}^{(H^i \times W^i) \times C^i}$ and $F_a^i \in \mathbb{R}^{(H^i \times W^i) \times C^i}$, respectively.  Further, following the methodologies of~\cite{choi2021robustnet,peng2021global}, we compute the covariance matrices using instance normalized features~\cite{ulyanov2016instance}, which ensures consistent scaling across features. The feature maps are normalized and flattened into $\bar{F}$ and  $\bar{F_a} \in \mathbb{R}^{(HW)\times C}$, which are given by:
\begin{align}
\bar{F} = \frac{\left(F - \mu(F)\right)}{\sigma(F)},
\bar{F_a} = \frac{\left(F_a - \mu(F_a)\right)}{\sigma(F_a)},
\end{align}
where $\mu(\cdot) \in \mathbb{R}^C$ and  $\sigma(\cdot) \in \mathbb{R}^C$ denote the mean and standard deviation of the features.
Utilizing the normalized features, we evaluate the  covariance matrices for the original and augmented image features as:
\begin{align}
\Sigma_{x,x}^i &= \bar{F^i}^T \cdot \bar{F^i}, & \Sigma_{x_a,x_a}^i  &= \bar{F_a^i}^T \cdot \bar{F_a^i}.
\end{align}
The CML is then formulated to align these covariance matrices, ensuring that the network maintains consistency in the presence of style variations. The CML is defined as:
\begin{equation}
\label{eq:cm}
\mathcal{L}_{CM} = \sum_{i=1}^{n_e} \|\Sigma_{x,x}^i - \Sigma_{x_a,x_a}^i\|_2,
\end{equation}
where $n_e$ denotes the number of blocks of the encoder.

\subsubsection*{Cross-covariance Loss}
While the CML effectively aligns the internal distributions of features within the same image, it does not fully account for the direct correlations across paired images. To complement this, we introduce CCL, which aims to encode the consistent content information of an image pair $(x, x_a)$ by utilizing the cross-covariance of the image pair. Given the normalized feature pair $(\bar{F^i}, \bar{F_a^i})$, the cross-covariance of the feature pair can be expressed as:
\begin{equation}
\Sigma_{x,x_a}^i = \bar{F^i}^T \cdot \bar{F_a^i}.
\end{equation}
The cross-covariance is expected to exhibit an identity matrix, as the feature pair should contain identical information. Nonetheless, the proposed CCL converges only the diagonal component of the covariance matrix to one.
This is to prevent the drawbacks of the existing FN methods~\cite{choi2021robustnet,peng2022semantic} from removing content information. The CCL function is thus defined as:
\begin{equation}
\label{eq:cc}
\mathcal{L}_{CC} = \sum_{i=1}^{n_e} \|\text{diag}(\Sigma_{x,x_a}^i) - \mathbbm{1}\|_2,
\end{equation}
where $\text{diag}(\Sigma_c) \in \mathbb{R}^C$ denotes the column vector comprising  diagonal elements of $\Sigma_{x,x_a}^i$ and $\mathbbm{1} \in \mathbb{R}^C$ denotes the one vector.


\begin{figure*}[ht]
  \centering
  \includegraphics[width=0.95\linewidth]{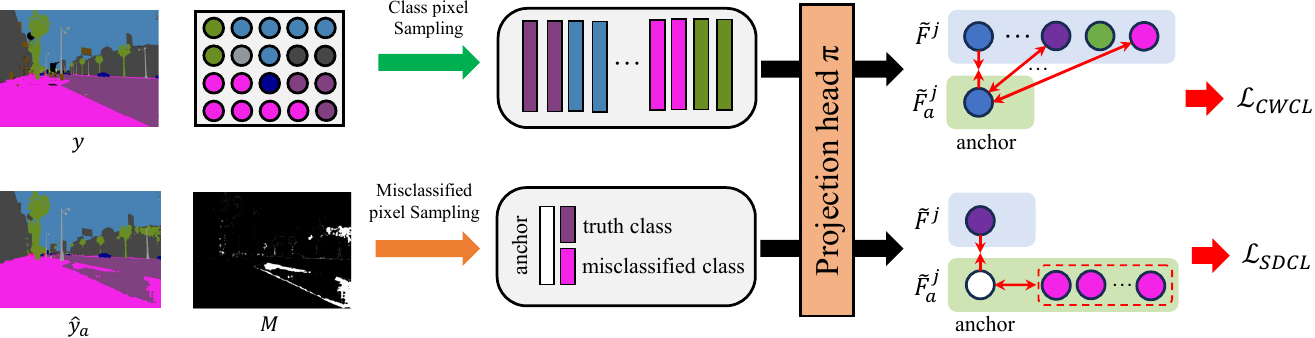}
  \caption{Illustration of semantic consistency contrastive learning: The mask $M$ represents the error mask derived from the augmented segmentation map. CWCL conducts contrastive learning by sampling per segmentation class and SDCL conducts contrastive learning based on the $M$. Both methods share a projection head $\pi$ for the semantic representation.}
  \label{fig:contrastive}
\end{figure*}

\subsection{Semantic Consistence Contrastive Learning}
While the encoder focuses on generating style-blinded features, the decoder aims to improve the robustness of the segmentation prediction against domain shifts. For the decoder, we employ a contrastive learning approach, which has demonstrated effectiveness in extracting discriminative features~\cite{chen2020simple}. Specifically, we utilize the InfoNCE loss~\cite{oord2018representation}, which is formulated as:
\begin{align}
\begin{aligned}
\mathcal{L}_{IN}(a,p,n) = -\log \left( \frac{e^{(a\cdot p/\tau)}}{e^{(a\cdot p/\tau)} + \sum_{n}^{N^-} e^{(a\cdot n/\tau)}} \right)
\end{aligned}
\end{align}
where $a$, $p$, $n$, and $N^-$ denote anchor, positive sample, negative sample, and negative sample set, respectively.

To achieve consistent feature representation in DGSS across various styles, we introduce semantic consistency contrastive learning. Specifically, the anchor is derived from the augmented image $x_a$, while the positive sample is extracted from the corresponding pixel of the original image $x$. Our method consists of two main components based on the negative sample as shown in Fig.~\ref{fig:contrastive}: class-wise contrastive learning and semantic disentanglement contrastive learning.

\subsubsection*{Class-wise Contrastive Learning}
Our CWCL aims to build a discriminative embedding space for each segmentation class using different classes of the original image as negatives.
Given an image pair $(x,x_a)$, the features from the $j^{th}$ block of the decoder at pixel position $(m,n)$ are denoted as $F_{(m,n)}^j$ and $F_{a,(m,n)}^j \in \mathbb{R}^{(1 \times 1) \times C^j}$, where $C^j$ indicates the channel length of the feature. As mentioned above, we take $F_{a,(m,n)}^j$ as the anchor and $F_{(m,n)}^j$ as the positive sample, since they represent the same content at the corresponding spatial location. To obtain the negative samples from $F^j$, we leverage the resized segmentation class label $y^j\in \mathbb{R}^{(H^j \times W^j) \times C}$ to collect the different class samples. The samples are passed through the projection head, denoted as $\pi$, resulting in the projected features $\Tilde{F}$ and $\Tilde{F}_a$. We define our CWCL as:
\begin{align}
\label{eq:cwcl}
\mathcal{L}_{CWCL}= \sum_{j}^{n_d}\mathcal{L}_{IN} \left(\Tilde{F}_{a,(m,n)}^j, \Tilde{F}_{(m,n)}^j, \Tilde{F}_{(p,q)}^j \right) \\
\nonumber\textrm{where} \quad (p,q)\in \{(p,q) \in P | y_{(p,q)}^j \neq y_{(m,n)}^j\}
\end{align}
The set $P$ represents all pixel positions in feature $F^j$, with dimensions $H^j \times W^j$ corresponding to the height and width. Additionally, $n_d$ denotes the total number of blocks in the decoder.

\begin{figure}[t]
  \centering
  \includegraphics[width=0.8\linewidth]{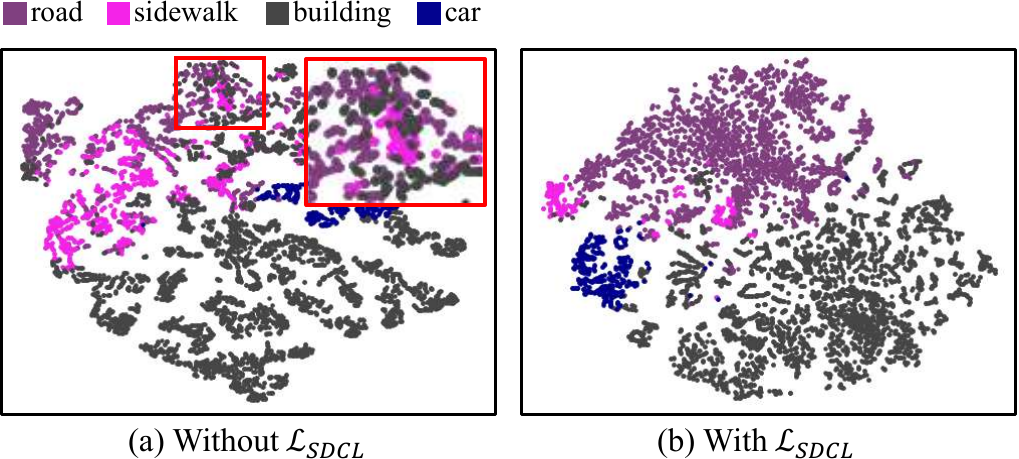}
  \caption{t-SNE~\cite{van2008visualizing} visualization comparing scenarios with and without $\mathcal{L}_{SCCL}$. In (b), the application of SCCL results in a clear separation between the sidewalk (pink), the road (purple), and the building (gray).}
  \label{fig:tsne}
\end{figure}

\begin{table*}[h]
\resizebox{\textwidth}{!}{%
\begin{tabular}{c|l|c|c|c|c|c|c||c|c|c|c}
\hline
\multirow{2}{*}{Backbone} & \multicolumn{1}{c|}{\multirow{2}{*}{Methods}} & \multicolumn{2}{c|}{External} & \multicolumn{4}{c||}{Trained on GTAV (G)} & \multicolumn{4}{c}{Trained on Cityscapes (C)} \\
 & \multicolumn{1}{c|}{} &\multicolumn{1}{c}{Dataset} & \multicolumn{1}{c|}{Module} & \multicolumn{1}{c}{C} & \multicolumn{1}{c}{B} & \multicolumn{1}{c}{M} & S & \multicolumn{1}{c}{B} & \multicolumn{1}{c}{M} & \multicolumn{1}{c}{S} & \multicolumn{1}{c}{G} \\ \hline \hline
\multirow{10}{*}{ResNet50~\cite{he2016deep}} & Baseline~\cite{chen2017rethinking} & - & - & 28.95 & 25.14 & 28.18 & 26.23 & 44.96 & 51.68 & 23.29 & 42.55 \\
 & IBN-Net~\cite{pan2018two} & - & - & 33.85 & 32.30 & 37.75 & 27.90 & 48.56 & 57.04 & 26.14 & 45.06  \\
 & RobustNet~\cite{choi2021robustnet} &- &-  & 37.31 & 35.20 & 40.33 & 28.30 & 50.73 & 58.64 & 26.20 & 45.00 \\
  & SiamDoGe~\cite{wu2022siamdoge} &-  & - & 42.96 & 37.54 & 40.64 & 28.34 & 51.53 & 59.00 & 26.67 & 45.08 \\
  & DIRL~\cite{xu2022dirl} & - & \checkmark & 41.04 & 39.15 & 41.60 & - & 51.80 & - & 26.50 &  46.52 \\
 & WildNet~\cite{lee2022wildnet} & \checkmark & - & 44.62 & 38.42 & 46.09 & \underline{31.34} & 50.94 & 58.79 & 27.95  & 47.01 \\
 & SANSAW~\cite{peng2022semantic} & - & \checkmark & 39.75 & 37.34 & 41.86 & 30.79 & \textbf{52.95} & \underline{59.81} & \underline{28.32}  & \underline{47.28} \\
 & SPC~\cite{huang2023style} &-  & \checkmark & 44.10 & \underline{40.46} & 45.51 & - & - & - & - & - \\
 & DPCL~\cite{yang2023generalized} &- & \checkmark & \underline{44.74} & 40.59 & \underline{46.33} & 30.81 & {50.97} & 58.59 & 25.85 & 46.00 \\
 & Ours & - & - & \textbf{45.72} & \textbf{41.32} & \textbf{47.08} & \textbf{31.39} & \underline{51.84} & \textbf{60.18} & \textbf{28.51} & \textbf{47.97} \\ \hline
\multirow{7}{*}{ShuffleNetV2~\cite{ma2018shufflenet}} 
& Baseline~\cite{chen2017rethinking} &- &- & 25.56 & 22.17 & 28.60 & 23.33 & 36.84 & 43.13 & 21.56 & 36.95 \\
 & IBN-Net~\cite{pan2018two} & -& -& 27.10 & 31.82 & 34.89 & \underline{25.56} & 41.89 & 46.35 & 22.99 & 40.91 \\
 & RobustNet~\cite{choi2021robustnet} & - & - & 30.98 & 32.06 & 35.31 & 24.31 & 41.94 & 46.97 & 22.82 & 40.17 \\
  & SiamDoGe~\cite{wu2022siamdoge} & - & -  & 34.40 & 34.23 & 35.87 & 21.95 & 42.61 & 47.48 & 23.13 & 40.93 \\
  & DIRL~\cite{xu2022dirl} & - & \checkmark & 31.88 & 32.57 & 36.12 & - & 42.55 & - & \textbf{23.74} & \textbf{41.23} \\
 & DPCL~\cite{yang2023generalized} & - & \checkmark & \underline{36.66} & \underline{34.35} & \underline{39.92} & 22.66 & \underline{43.90} & \underline{48.95} & 22.47 & {41.07} \\
 & Ours &- & -& \textbf{38.56} & \textbf{34.51} & \textbf{40.11} & \textbf{25.64} & \textbf{44.22} & \textbf{49.69} & \underline{23.54} & \underline{41.10} \\ \hline
\end{tabular}%
}
\caption{Quantitative comparison of mIoU (\%) between DGSS methods. External dataset denotes the necessity of an auxiliary dataset during training and External module denotes the requirement of an additional module during inference. G, C, B, M, and S denote GTAV, Cityscapes, BDD100K, Mapillary, and SYNTHIA, respectively. The best and second-best results are \textbf{bolded} and \underline{underlined}, respectively.}
\label{tab:result}
\end{table*}

\subsubsection*{Semantic Disentanglement Contrastive Learning}
Domain shifts can lead to the entanglement of similar classes, causing the model to misclassify, as illustrated in Fig.~\ref{fig:tsne}. To mitigate this issue, we introduce the SDCL, specifically designed to disentangle the feature $x_a$ that has been misclassified, making it closer to the correct class and far from the misclassified class to achieve effective disentanglement. To further ensure consistent feature space and capture the semantic meaning, we share the projection head $\pi$ used in the CWCL loss.
Given the predicted segmentation map of the augmented image, represented as $\hat{y}_a = \varphi(x_a)$, we resize it to $\hat{y}_{a,(m,n)}^j \in \mathbb{R}^{(H^j \times W^j) \times C}$. Similarly, $y_{(m,n)}^j$ represents the ground truth segmentation map. Using these segmentation maps, we set the anchor at positions where $\hat{y}_{a,(m,n)}^j \neq y_{(m,n)}^j$. Negative samples are selected from the augmented image features corresponding to the anchor's misclassified class. The samples go through the projection head $\pi$. Our SDCL loss is defined as follows:
\begin{align}
\label{eq:sdcl}
\mathcal{L}_{SDCL} = \sum_{j}^{n_d}\mathcal{L}_{IN} \left(\Tilde{F}_{a,(m,n)}^j, \Tilde{F}_{(m,n)}^j, \Tilde{F}_{a,(r,s)}^j \right) \\ 
\nonumber\textrm{where} \quad (r,s) \in \{(r,s) \in P | y_{(r,s)}^j = \hat{y}_{(m,n)}^j\}
\end{align}

Finally, combining the cross-entropy segmentation loss $\mathcal{L}_{CE}$ with other loss components, the total is defined as:
\begin{align}
\begin{aligned}
    \mathcal{L}_{\text{total}} = \mathcal{L}_{CE} &+ \omega_1 \mathcal{L}_{CM} + \omega_2 \mathcal{L}_{CC} \\
    &+ \omega_3 \mathcal{L}_{CWCL}+ \omega_4 \mathcal{L}_{SDCL}
\end{aligned}
\label{eq:total}
\end{align}
where $\omega_1$, $\omega_2$, $\omega_3$, and $\omega_4$ denote the weighting factor of each loss functions.

\section{Experiment}
\label{sec:exp}

In this section, we describe the implementation details, the experimental setup for comparison with existing DGSS methods, and the ablation study conducted to further validate the effectiveness of our approach.

\subsection{Implementation Details}
\label{sec:implmentation}
We adopt DeepLabV3+~\cite{chen2017rethinking} for the segmentation architecture and use ResNet-50~\cite{he2016deep}, ShuffleNetV2~\cite{ma2018shufflenet}, and MoblieNetV2~\cite{sandler2018mobilenetv2} as the backbone network of the segmentation network. The model is trained for 40K iterations with a batch size of 8 using the SGD optimizer, which has a momentum of 0.9 and a weight decay of 5e-4. We employ a polynomial learning rate schedule with an initial rate of 1e-2 and a power of 0.9. For the simulation of domain shift, we augment the image $x_a$ using strong color jittering transformation similar to~\cite{choi2021robustnet}. The weighting parameters of (\ref{eq:total}), $\omega_1$, $\omega_2$, $\omega_3$ and $\omega_4$, are set as 0.2, 0.2, 0.3, and 0.3 respectively.

\subsection{Datasets}
We use two synthetic datasets (GTA~\cite{richter2016playing} and SYNTHIA~\cite{ros2016synthia}), and three real-world datasets (Cityscapes~\cite{cordts2016cityscapes}, BDD-100K~\cite{yu2020bdd100k}, and Mapillary~\cite{neuhold2017mapillary}) for the experiment. All segmentation labels are evaluated based on 19 object categories.

\noindent\textbf{GTAV (G)}~\cite{richter2016playing} is a large-scale dataset generated from the Grand Theft Auto V (GTAV)  game engine.  It comprises 24,966 images, split into 12,403 for training, 6,382 for validation, and 6,181 for testing with a resolution of 1914$\times$1052.

\noindent\textbf{SYNTHIA (S)}~\cite{ros2016synthia} is a virtual, photo-realistic urban scene dataset comprising 9,400 images with a resolution of 960$\times$720. Among these, 2,820 images are designated for evaluation.

\noindent \textbf{Cityscapes (C)}~\cite{cordts2016cityscapes} is a large-scale urban scene dataset captured from 50 cities, primarily in Germany. Particularly, it contains  5,000 high-resolution images with a resolution of 2048$\times$1024. The dataset is divided into 2,975 images for training, 500 for validation, and 1,525 for testing.

\noindent \textbf{BDD-100K (B)}~\cite{yu2020bdd100k} is another real-world urban scene dataset that contains more diverse 10000 urban driving scene images with a resolution of 1280$\times$720. Specifically, the validation split (1,000 images) is used for evaluation.

\noindent \textbf{Mapillary (M)}~\cite{neuhold2017mapillary} contains 25,000 images with a minimum resolution of 1920$\times$1080, collected  from various locations worldwide. Specifically, the validation split of 2,000 images is used for evaluation.

\subsection{Comparison with DGSS methods}
We compare our methods with other state-of-the-art DGSS methods: Baseline (DeepLabV3+~\cite{chen2017rethinking} trained on the source domain), IBN-Net~\cite{pan2018two}, RobustNet~\cite{choi2021robustnet}, SiamDoGe~\cite{wu2022siamdoge}, DIRL~\cite{xu2022dirl}, WildNet~\cite{lee2022wildnet}, SANSAW~\cite{peng2022semantic}, SPC~\cite{huang2023style}, and  DPCL~\cite{yang2023generalized}. To evaluate the generalization ability of the model on arbitrary unseen domains, we conduct the experiment on two scenarios: i) trained on GTAV, tested on Cityscapes, BDD-100K, and Mapillary, and ii) trained on Cityscapes, tested on BDD-100K, Mapillary, and SYNTHIA. The quantitative results are computed with mean intersection over union (mIoU).  Additionally, we compared the method trained on the backbone of ResNet-50~\cite{he2016deep}, ShuffleNetV2~\cite{ma2018shufflenet}, and MoblieNet~\cite{sandler2018mobilenetv2}, pre-trained on ImageNet~\cite{deng2009imagenet}.

\begin{table}[]
\resizebox{\columnwidth}{!}{%
\begin{tabular}{l|c|ccc||c}
\hline
\multicolumn{1}{c|}{\multirow{2}{*}{Methods}} & \multirow{2}{*}{\begin{tabular}[c]{@{}c@{}}External\\ Module\end{tabular}} & \multicolumn{4}{c}{Trained on GTAV (G)} \\
\multicolumn{1}{c|}{} &  & C & B & M & Mean \\ \hline
Baseline~\cite{chen2017rethinking} &  & 25.94 & 25.73 & 26.45 & 26.04 \\
IBN-Net~\cite{pan2018two} &  & 30.14 & 27.66 & 27.07 & 28.29 \\
RobustNet~\cite{choi2021robustnet} &  & 30.86 & 30.05 & 30.67 & 30.52 \\
SiamDoGe~\cite{wu2022siamdoge} &  & 34.15 & 34.50 & 32.34 & 33.67 \\
DIRL~\cite{xu2022dirl} & \checkmark & 34.67 & 32.78 & 34.31 & 33.92 \\
DPCL~\cite{yang2023generalized} & \checkmark & \underline{37.57} & \underline{35.45} & \underline{40.30} & \underline{37.77} \\
Ours &  & \textbf{37.66} & \textbf{36.10} & \textbf{40.40} & \textbf{38.05} \\ \hline
\end{tabular}%
}
\caption{Quantitative comparison of mIoU (\%) using MobileNetV2~\cite{sandler2018mobilenetv2} backbone trained on the GTAV dataset.}
\label{tab:mobliegta}
\end{table}

\subsubsection*{Quantitative and Qualitative Results}
Table~\ref{tab:result} summarizes the quantitative results. Our method outperforms all other methods when trained on GTAV, using ResNet-50 as the backbone. When compared with FN methods that remove domain-specific styles, we demonstrate that our approach minimizes the loss of content information. We also show that our method effectively shows generalization ability when trained on Cityscapes. We further evaluate our methods with different backbones, showing the wide applicability of our method. When trained with ShuffleNetV2, our method achieves the first or second-best performance among unseen target domains. Table~\ref{tab:mobliegta} shows the results of our method trained on GTAV with MobileNetV2, demonstrating the superiority of our method.

\begin{figure*}
  \centering
  \includegraphics[width=\linewidth]{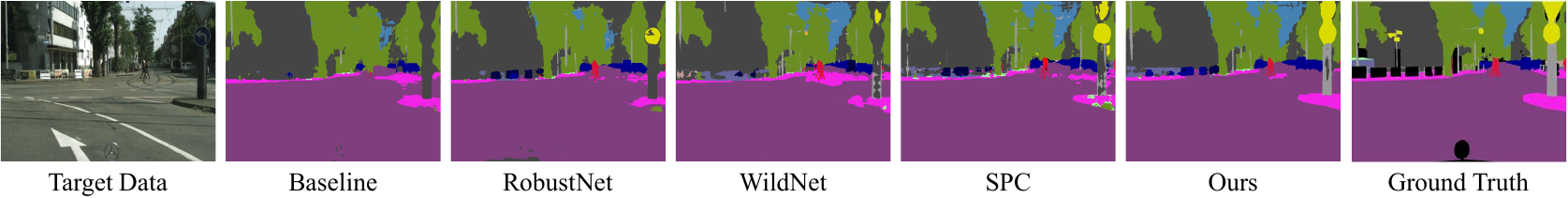}
  \caption{Qualitative comparison between DGSS methods trained on GTAV (G) and tested on unseen target domains of Cityscapes (C) using DeeplabV3+ with ResNet50 backbone.}
  \label{fig:result}
\end{figure*}

For qualitative evaluation, we compare the visual result between DGSS methods and ours. As depicted in Fig.~\ref{fig:result}, our method demonstrates superior results compared to other approaches, particularly in its overall prediction accuracy. Notably, our proposed techniques enable distinct prediction of features such as on road and sidewalk, yielding clearer segmentation boundaries. Please refer to the supplementary material for more qualitative results.

\begin{table}[]
\resizebox{\linewidth}{!}{%
\begin{tabular}{l|c|ccc}
\hline
Methods & \begin{tabular}[c]{@{}c@{}}Externel\\ Module\end{tabular} & Params (M) & GFLOPS & Time (ms) \\ \hline \hline 
Baseline~\cite{he2016deep} &  & 45.08 & 277.77 & 10.01 \\
SANSAW~\cite{peng2022semantic} & \checkmark & 25.63 & 421.86 & 68.96 \\
SPC~\cite{huang2023style} & \checkmark & 45.22 & 286.09 & 12.24 \\
DIRL~\cite{xu2022dirl} & \checkmark & 45.41 & 278.11 & 11.69 \\
DPCL~\cite{yang2023generalized} & \checkmark & 56.46 & 1188.64 & 823.78 \\
Ours &  & 45.08 & 277.78 & 10.03 \\ \hline
\end{tabular}%
}
\caption{Computational cost comparison conducted using DeepLabV3+ with a ResNet-50 backbone on an NVIDIA Tesla V100 GPU with an image resolution of $2048 \times 1024$. Inference time is averaged over 400 trials.}
\label{tab:cost}
\end{table}

\subsubsection*{Computational cost analysis}
To confirm that our approach does not incur additional computational overhead, we provide the number of parameters, GFLOPS, and average inference time of each method. As detailed in Table~\ref{tab:cost}, our method operates comparably to baseline models by learning features intrinsically without adopting a separate module.


\begin{table}[]
\resizebox{\linewidth}{!}{%
\begin{tabular}{P{1cm}P{1cm}|P{1cm}P{1cm}|ccc}
\hline
$\mathcal{L}_{CM} $ & $\mathcal{L}_{CC}$ & $\mathcal{L}_{CWCL}$ & $\mathcal{L}_{SDCL}$ & C & B & M  \\ \hline \hline
 &   &   &  & 28.95 & 25.14 & 28.18 \\ 
\checkmark &  &  &  & 38.08  & 36.65 &40.62  \\
\checkmark & \checkmark& & & 40.42& 37.81 & 43.91 \\
 &  &  \checkmark  & & 42.03 & 38.27 & 44.02  \\
& &  \checkmark  &  \checkmark & 43.16 & \underline{38.59} & \underline{45.38}  \\
\checkmark  &  \checkmark  & \checkmark  &  & \underline{43.17}  & 38.23 & 44.84  \\
  \checkmark      &  \checkmark       &    \checkmark      & \checkmark      & \textbf{45.72}  & \textbf{41.32}  &  \textbf{47.08}  \\ \hline
\end{tabular}}
\caption{Ablation study on proposed losses. The experiments were conducted using DeepLabV3+ with ResNet-50 backbone, trained on the GTAV dataset. The losses are detailed in $\mathcal{L}_{CM}$: (\ref{eq:cm}), $\mathcal{L}_{CC}$: (\ref{eq:cc}), $\mathcal{L}_{CWCL}$: (\ref{eq:cwcl}), $\mathcal{L}_{SDCL}$: (\ref{eq:sdcl})}
\label{tab:ablation}
\end{table}

\subsection{Ablation Studies}
In this subsection, we conducted a series of ablation studies to demonstrate the individual contribution and effectiveness of each component within our method. Specifically, we investigate the impact of the following components: $\mathcal{L}_{CM}$, $\mathcal{L}_{CC}$, $\mathcal{L}_{CWCL}$, $\mathcal{L}_{SDCL}$. For the study, we use a scenario where the DeepLabV3+ model with backbone ResNet-50 model is trained on GTA and tested on Cityscapes, BDD-100K, and Mapillary.

Table~\ref{tab:ablation} presents the impact of various proposed losses on domain generalization performance. Specifically, the baseline model, trained solely with cross-entropy loss, exhibits suboptimal performance on target domains because of overfitting to the source domain. Conversely, the integration of any proposed loss mechanisms leads to a marked enhancement in performance. More specifically, the incorporation of the covariance alignment ($\mathcal{L}_{CM}$, $\mathcal{L}_{CC}$) shows its efficacy in preserving essential content information by correlating features of paired images. The differential impact of the semantic consistency constrastive learning ($\mathcal{L}_{CWCL}$, $\mathcal{L}_{SDCL}$) is also evident, as it significantly aids in disentangling features of similar classes, thereby constructing a more robust embedding space.

\begin{figure}[t]
  \centering
  \includegraphics[width=0.8\linewidth]{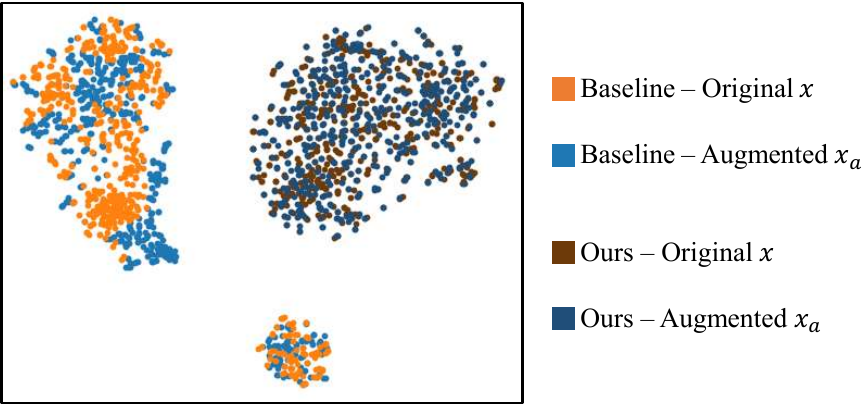}
  \caption{t-SNE~\cite{van2008visualizing} visualization comparing the covariance with and without $\mathcal{L}_{CML}$.}
  \label{fig:cml}
\end{figure}

\noindent \textbf{Covariance Matching Loss.} Fig.~\ref{fig:cml} presents t-SNE plots of covariances for original and augmented images, before and after the application of CML. The baseline network perceives original and augmented images differently from a style perspective. However, after applying CML, the distribution becomes more intermixed, indicating that our proposed CML effectively ensures similar recognition of different style images.

\noindent \textbf{Calcuation of CCL.} Table~\ref{tab:cc} demonstrates that our proposed cross-covariance method, which converges the diagonal components to 1, yields superior performance. As mentioned before,  removing non-diagonal components, which contain content information actually degrades performance.

\noindent \textbf{Sampling number in CWCL.} Table~\ref{tab:numcwcl} and Table~\ref{tab:negcwcl} show the impact of varying the number of classes sampled per image and the number of samples per class in CWCL, respectively. As shown in Table~\ref{tab:numcwcl}, the performance improves with an increase in the diversity of classes sampled in CWCL. This suggests that contrasting a broader array of classes enhances the model's discriminative capability. Furthermore, Table~\ref{tab:negcwcl} demonstrates that a balanced number of negative samples per class leads to optimal performance.

\noindent \textbf{Projection Head for SDCL.} The influence of different project head configurations on the SDCL is investigated. We experimented with three distinct approaches: individual projection head, copying the weights of CWCL's (stop gradient), and shared projection head of CWCL. As demonstrated in Table~\ref{tab:projhead}, sharing the projection head yielded the most superior results. The results indicate that SDCL not only relies on the semantic information from CWCL for effective disentanglement of similar classes but also enhances the embedding space learned by CWCL.

\begin{table}[t]
\centering

\resizebox{\linewidth}{!}{%

\begin{subtable}[t]{.65\linewidth}
\centering
\begin{tabular}{cccc}
\hline
\multicolumn{4}{c}{Cross-covariance loss} \\ \hline \hline
\multicolumn{1}{c|}{Method} & C & B & M \\ \hline \hline
\multicolumn{1}{c|}{Whitening} & 38.68 & 36.91 & 42.12 \\
\multicolumn{1}{c|}{Diagonal} & 40.42 & 37.81 & 43.91 \\ \hline
\end{tabular}
\caption{}
\label{tab:cc}
\end{subtable}

\begin{subtable}[t]{.6\linewidth}
\centering
\begin{tabular}{cccc}
\hline
\multicolumn{4}{c}{\# of classes} \\ \hline \hline
\multicolumn{1}{c|}{\#} & C & B & M \\ \hline \hline
\multicolumn{1}{c|}{10} & 45.57 & 38.88 & 46.37 \\
\multicolumn{1}{c|}{15} & 45.72 & 41.32 & 47.08 \\ 
 \hline
\end{tabular}
\caption{}
\label{tab:numcwcl}
\end{subtable}%
}
\vspace{0.2cm}

\resizebox{\linewidth}{!}{%

\begin{subtable}[t]{.6\linewidth}
\centering
\begin{tabular}{cccc}
\hline
\multicolumn{4}{c}{\# of negative samples} \\ \hline \hline
\multicolumn{1}{c|}{\#} & C & B & M \\ \hline \hline
\multicolumn{1}{c|}{10} & 44.76 & 38.21 & 46.46 \\
\multicolumn{1}{c|}{50} & 45.72 & 41.32 & 47.08 \\
\multicolumn{1}{c|}{100} & 44.44 & 39.14 & 46.29 \\ \hline
\end{tabular}
\caption{}
\label{tab:negcwcl}
\end{subtable}

\begin{subtable}[t]{.73\linewidth}
\centering
\begin{tabular}{cccc}
\hline
\multicolumn{4}{c}{Projection Head of SDCL} \\ \hline \hline
\multicolumn{1}{c|}{MLP} & C & B & M \\ \hline \hline
\multicolumn{1}{c|}{Individual} & 44.91 & 38.45 & 46.31 \\
\multicolumn{1}{c|}{Shared (SG)} & 44.03 & 38.15 & 46.64 \\ 
\multicolumn{1}{c|}{Shared} & 45.72 & 41.32 & 47.08 \\ \hline
\end{tabular}
\caption{}
\label{tab:projhead}
\end{subtable}
}

\caption{Ablation studies. (a) Calculation of $\mathcal{L}_{CC}$. (b) Number of classes for $\mathcal{L}_{CWCL}$. (c) Number of negative samples for $\mathcal{L}_{CWCL}$. (d) Projection head of $\mathcal{L}_{SDCL}$. ``SG'' indicates stop gradient.}
\label{table:overall_label}
\end{table}
\section{Conclusion}
\label{sec:con}

In this paper, we propose a novel BlindNet with covariance alignment and semantic consistency contrastive learning. By introducing covariance alignment, our method effectively addresses style variations, ensuring the extraction of features that are consistent across different styles. Furthermore, with the proposed semantic consistency contrastive learning, we not only facilitate the extraction of discriminative features but also enhance the generalization capabilities of the model in semantic segmentation predictions. Comprehensive experimental results validate the effectiveness of our approach, demonstrating its ability to generalize across multiple unseen target domains without requiring auxiliary domains or additional modules. Our future work will be improving and stabilizing the covariance alignment method.

\vspace{0.1cm}

\noindent\textbf{Acknowledgement.} This work was supported in part by the Basic Science Research Program through National Research Foundation of Korea (NRF) (Grants No. NRF-2022R1F1A1073543), the MSIT(Ministry of Science and ICT), Korea, under the ICAN(ICT Challenge and Advanced Network of HRD) support program(RS-2022-00156385) supervised by the IITP(Institute for Information \& Communications Technology Planning \& Evaluation), and Innovative Human Resource Development for Local Intellectualization program through the Institute of Information \& Communications Technology Planning \& Evaluation(IITP) grant funded by the Korea government(MSIT)(IITP-2024-00156287).
\clearpage
{
    \small
    \bibliographystyle{ieeenat_fullname}
    \bibliography{main}

\begin{thebibliography}{62}
\providecommand{\natexlab}[1]{#1}
\providecommand{\url}[1]{\texttt{#1}}
\expandafter\ifx\csname urlstyle\endcsname\relax
  \providecommand{\doi}[1]{doi: #1}\else
  \providecommand{\doi}{doi: \begingroup \urlstyle{rm}\Url}\fi

\bibitem[Bartoccioni et~al.(2023)Bartoccioni, Zablocki, Bursuc, P{\'e}rez,
  Cord, and Alahari]{bartoccioni2023lara}
Florent Bartoccioni, {\'E}loi Zablocki, Andrei Bursuc, Patrick P{\'e}rez,
  Matthieu Cord, and Karteek Alahari.
\newblock Lara: Latents and rays for multi-camera bird’s-eye-view semantic
  segmentation.
\newblock In \emph{Conference on Robot Learning}, pages 1663--1672. PMLR, 2023.

\bibitem[Caron et~al.(2020)Caron, Misra, Mairal, Goyal, Bojanowski, and
  Joulin]{caron2020unsupervised}
Mathilde Caron, Ishan Misra, Julien Mairal, Priya Goyal, Piotr Bojanowski, and
  Armand Joulin.
\newblock Unsupervised learning of visual features by contrasting cluster
  assignments.
\newblock \emph{Advances in neural information processing systems},
  33:\penalty0 9912--9924, 2020.

\bibitem[Chen et~al.(2017)Chen, Papandreou, Schroff, and
  Adam]{chen2017rethinking}
Liang-Chieh Chen, George Papandreou, Florian Schroff, and Hartwig Adam.
\newblock Rethinking atrous convolution for semantic image segmentation.
\newblock \emph{arXiv preprint arXiv:1706.05587}, 2017.

\bibitem[Chen et~al.(2020)Chen, Kornblith, Norouzi, and Hinton]{chen2020simple}
Ting Chen, Simon Kornblith, Mohammad Norouzi, and Geoffrey Hinton.
\newblock A simple framework for contrastive learning of visual
  representations.
\newblock In \emph{International conference on machine learning}, pages
  1597--1607. PMLR, 2020.

\bibitem[Cheng et~al.(2022)Cheng, Misra, Schwing, Kirillov, and
  Girdhar]{cheng2022masked}
Bowen Cheng, Ishan Misra, Alexander~G Schwing, Alexander Kirillov, and Rohit
  Girdhar.
\newblock Masked-attention mask transformer for universal image segmentation.
\newblock In \emph{Proceedings of the IEEE/CVF conference on computer vision
  and pattern recognition}, pages 1290--1299, 2022.

\bibitem[Cho et~al.(2019)Cho, Choi, Park, Shin, and Choo]{cho2019image}
Wonwoong Cho, Sungha Choi, David~Keetae Park, Inkyu Shin, and Jaegul Choo.
\newblock Image-to-image translation via group-wise deep whitening-and-coloring
  transformation.
\newblock In \emph{Proceedings of the IEEE/CVF Conference on Computer Vision
  and Pattern Recognition}, pages 10639--10647, 2019.

\bibitem[Choi et~al.(2021)Choi, Jung, Yun, Kim, Kim, and
  Choo]{choi2021robustnet}
Sungha Choi, Sanghun Jung, Huiwon Yun, Joanne~T Kim, Seungryong Kim, and Jaegul
  Choo.
\newblock Robustnet: Improving domain generalization in urban-scene
  segmentation via instance selective whitening.
\newblock In \emph{Proceedings of the IEEE/CVF Conference on Computer Vision
  and Pattern Recognition}, pages 11580--11590, 2021.

\bibitem[Cordts et~al.(2016)Cordts, Omran, Ramos, Rehfeld, Enzweiler, Benenson,
  Franke, Roth, and Schiele]{cordts2016cityscapes}
Marius Cordts, Mohamed Omran, Sebastian Ramos, Timo Rehfeld, Markus Enzweiler,
  Rodrigo Benenson, Uwe Franke, Stefan Roth, and Bernt Schiele.
\newblock The cityscapes dataset for semantic urban scene understanding.
\newblock In \emph{Proceedings of the IEEE conference on computer vision and
  pattern recognition}, pages 3213--3223, 2016.

\bibitem[Deng et~al.(2009)Deng, Dong, Socher, Li, Li, and
  Fei-Fei]{deng2009imagenet}
Jia Deng, Wei Dong, Richard Socher, Li-Jia Li, Kai Li, and Li Fei-Fei.
\newblock Imagenet: A large-scale hierarchical image database.
\newblock In \emph{2009 IEEE conference on computer vision and pattern
  recognition}, pages 248--255. Ieee, 2009.

\bibitem[Gatys et~al.(2016)Gatys, Ecker, and Bethge]{gatys2016image}
Leon~A Gatys, Alexander~S Ecker, and Matthias Bethge.
\newblock Image style transfer using convolutional neural networks.
\newblock In \emph{Proceedings of the IEEE conference on computer vision and
  pattern recognition}, pages 2414--2423, 2016.

\bibitem[Goodfellow et~al.(2014)Goodfellow, Pouget-Abadie, Mirza, Xu,
  Warde-Farley, Ozair, Courville, and Bengio]{goodfellow2014generative}
Ian Goodfellow, Jean Pouget-Abadie, Mehdi Mirza, Bing Xu, David Warde-Farley,
  Sherjil Ozair, Aaron Courville, and Yoshua Bengio.
\newblock Generative adversarial nets.
\newblock \emph{Advances in neural information processing systems}, 27, 2014.

\bibitem[Grill et~al.(2020)Grill, Strub, Altch{\'e}, Tallec, Richemond,
  Buchatskaya, Doersch, Avila~Pires, Guo, Gheshlaghi~Azar,
  et~al.]{grill2020bootstrap}
Jean-Bastien Grill, Florian Strub, Florent Altch{\'e}, Corentin Tallec, Pierre
  Richemond, Elena Buchatskaya, Carl Doersch, Bernardo Avila~Pires, Zhaohan
  Guo, Mohammad Gheshlaghi~Azar, et~al.
\newblock Bootstrap your own latent-a new approach to self-supervised learning.
\newblock \emph{Advances in neural information processing systems},
  33:\penalty0 21271--21284, 2020.

\bibitem[He et~al.(2016)He, Zhang, Ren, and Sun]{he2016deep}
Kaiming He, Xiangyu Zhang, Shaoqing Ren, and Jian Sun.
\newblock Deep residual learning for image recognition.
\newblock In \emph{Proceedings of the IEEE conference on computer vision and
  pattern recognition}, pages 770--778, 2016.

\bibitem[He et~al.(2020)He, Fan, Wu, Xie, and Girshick]{he2020momentum}
Kaiming He, Haoqi Fan, Yuxin Wu, Saining Xie, and Ross Girshick.
\newblock Momentum contrast for unsupervised visual representation learning.
\newblock In \emph{Proceedings of the IEEE/CVF conference on computer vision
  and pattern recognition}, pages 9729--9738, 2020.

\bibitem[Hoffman et~al.(2018)Hoffman, Tzeng, Park, Zhu, Isola, Saenko, Efros,
  and Darrell]{hoffman2018cycada}
Judy Hoffman, Eric Tzeng, Taesung Park, Jun-Yan Zhu, Phillip Isola, Kate
  Saenko, Alexei Efros, and Trevor Darrell.
\newblock Cycada: Cycle-consistent adversarial domain adaptation.
\newblock In \emph{International conference on machine learning}, pages
  1989--1998. Pmlr, 2018.

\bibitem[Hoyer et~al.(2022)Hoyer, Dai, and Van~Gool]{hoyer2022daformer}
Lukas Hoyer, Dengxin Dai, and Luc Van~Gool.
\newblock Daformer: Improving network architectures and training strategies for
  domain-adaptive semantic segmentation.
\newblock In \emph{Proceedings of the IEEE/CVF Conference on Computer Vision
  and Pattern Recognition}, pages 9924--9935, 2022.

\bibitem[Hoyer et~al.(2023)Hoyer, Dai, Wang, Chen, and
  Van~Gool]{hoyer2023improving}
Lukas Hoyer, Dengxin Dai, Qin Wang, Yuhua Chen, and Luc Van~Gool.
\newblock Improving semi-supervised and domain-adaptive semantic segmentation
  with self-supervised depth estimation.
\newblock \emph{International Journal of Computer Vision}, pages 1--27, 2023.

\bibitem[Hu et~al.(2023)Hu, Yang, Chen, Li, Sima, Zhu, Chai, Du, Lin, Wang,
  et~al.]{hu2023planning}
Yihan Hu, Jiazhi Yang, Li Chen, Keyu Li, Chonghao Sima, Xizhou Zhu, Siqi Chai,
  Senyao Du, Tianwei Lin, Wenhai Wang, et~al.
\newblock Planning-oriented autonomous driving.
\newblock In \emph{Proceedings of the IEEE/CVF Conference on Computer Vision
  and Pattern Recognition}, pages 17853--17862, 2023.

\bibitem[Huang et~al.(2021)Huang, Guan, Xiao, and Lu]{huang2021fsdr}
Jiaxing Huang, Dayan Guan, Aoran Xiao, and Shijian Lu.
\newblock Fsdr: Frequency space domain randomization for domain generalization.
\newblock In \emph{Proceedings of the IEEE/CVF Conference on Computer Vision
  and Pattern Recognition}, pages 6891--6902, 2021.

\bibitem[Huang et~al.(2023)Huang, Chen, Li, Li, Li, Song, Yan, and
  Xiong]{huang2023style}
Wei Huang, Chang Chen, Yong Li, Jiacheng Li, Cheng Li, Fenglong Song, Youliang
  Yan, and Zhiwei Xiong.
\newblock Style projected clustering for domain generalized semantic
  segmentation.
\newblock In \emph{Proceedings of the IEEE/CVF Conference on Computer Vision
  and Pattern Recognition}, pages 3061--3071, 2023.

\bibitem[Huang and Belongie(2017)]{huang2017arbitrary}
Xun Huang and Serge Belongie.
\newblock Arbitrary style transfer in real-time with adaptive instance
  normalization.
\newblock In \emph{Proceedings of the IEEE international conference on computer
  vision}, pages 1501--1510, 2017.

\bibitem[Huang et~al.(2018)Huang, Liu, Belongie, and
  Kautz]{huang2018multimodal}
Xun Huang, Ming-Yu Liu, Serge Belongie, and Jan Kautz.
\newblock Multimodal unsupervised image-to-image translation.
\newblock In \emph{Proceedings of the European conference on computer vision
  (ECCV)}, pages 172--189, 2018.

\bibitem[Ioffe and Szegedy(2015)]{ioffe2015batch}
Sergey Ioffe and Christian Szegedy.
\newblock Batch normalization: Accelerating deep network training by reducing
  internal covariate shift.
\newblock In \emph{International conference on machine learning}, pages
  448--456. pmlr, 2015.

\bibitem[Lee et~al.(2022)Lee, Seong, Lee, and Kim]{lee2022wildnet}
Suhyeon Lee, Hongje Seong, Seongwon Lee, and Euntai Kim.
\newblock Wildnet: Learning domain generalized semantic segmentation from the
  wild.
\newblock In \emph{Proceedings of the IEEE/CVF Conference on Computer Vision
  and Pattern Recognition}, pages 9936--9946, 2022.

\bibitem[Li et~al.(2020)Li, Kang, Liu, Wei, and Yang]{li2020content}
Guangrui Li, Guoliang Kang, Wu Liu, Yunchao Wei, and Yi Yang.
\newblock Content-consistent matching for domain adaptive semantic
  segmentation.
\newblock In \emph{European conference on computer vision}, pages 440--456.
  Springer, 2020.

\bibitem[Li et~al.(2021)Li, Gao, Cao, Huang, Weng, Mi, Yu, Li, and
  Xia]{li2021progressive}
Lei Li, Ke Gao, Juan Cao, Ziyao Huang, Yepeng Weng, Xiaoyue Mi, Zhengze Yu,
  Xiaoya Li, and Boyang Xia.
\newblock Progressive domain expansion network for single domain
  generalization.
\newblock In \emph{Proceedings of the IEEE/CVF Conference on Computer Vision
  and Pattern Recognition}, pages 224--233, 2021.

\bibitem[Li et~al.(2017)Li, Fang, Yang, Wang, Lu, and Yang]{li2017universal}
Yijun Li, Chen Fang, Jimei Yang, Zhaowen Wang, Xin Lu, and Ming-Hsuan Yang.
\newblock Universal style transfer via feature transforms.
\newblock \emph{Advances in neural information processing systems}, 30, 2017.

\bibitem[Li et~al.(2018)Li, Tian, Gong, Liu, Liu, Zhang, and Tao]{li2018deep}
Ya Li, Xinmei Tian, Mingming Gong, Yajing Liu, Tongliang Liu, Kun Zhang, and
  Dacheng Tao.
\newblock Deep domain generalization via conditional invariant adversarial
  networks.
\newblock In \emph{Proceedings of the European conference on computer vision
  (ECCV)}, pages 624--639, 2018.

\bibitem[Li et~al.(2019)Li, Yuan, and Vasconcelos]{li2019bidirectional}
Yunsheng Li, Lu Yuan, and Nuno Vasconcelos.
\newblock Bidirectional learning for domain adaptation of semantic
  segmentation.
\newblock In \emph{Proceedings of the IEEE/CVF Conference on Computer Vision
  and Pattern Recognition}, pages 6936--6945, 2019.

\bibitem[Li and Hoiem(2017)]{li2017learning}
Zhizhong Li and Derek Hoiem.
\newblock Learning without forgetting.
\newblock \emph{IEEE transactions on pattern analysis and machine
  intelligence}, 40\penalty0 (12):\penalty0 2935--2947, 2017.

\bibitem[Luo(2017)]{luo2017learning}
Ping Luo.
\newblock Learning deep architectures via generalized whitened neural networks.
\newblock In \emph{International Conference on Machine Learning}, pages
  2238--2246. PMLR, 2017.

\bibitem[Ma et~al.(2018)Ma, Zhang, Zheng, and Sun]{ma2018shufflenet}
Ningning Ma, Xiangyu Zhang, Hai-Tao Zheng, and Jian Sun.
\newblock Shufflenet v2: Practical guidelines for efficient cnn architecture
  design.
\newblock In \emph{Proceedings of the European conference on computer vision
  (ECCV)}, pages 116--131, 2018.

\bibitem[Neuhold et~al.(2017)Neuhold, Ollmann, Rota~Bulo, and
  Kontschieder]{neuhold2017mapillary}
Gerhard Neuhold, Tobias Ollmann, Samuel Rota~Bulo, and Peter Kontschieder.
\newblock The mapillary vistas dataset for semantic understanding of street
  scenes.
\newblock In \emph{Proceedings of the IEEE international conference on computer
  vision}, pages 4990--4999, 2017.

\bibitem[Nilsson et~al.(2021)Nilsson, Pirinen, G{\"a}rtner, and
  Sminchisescu]{nilsson2021embodied}
David Nilsson, Aleksis Pirinen, Erik G{\"a}rtner, and Cristian Sminchisescu.
\newblock Embodied visual active learning for semantic segmentation.
\newblock In \emph{Proceedings of the AAAI Conference on Artificial
  Intelligence}, pages 2373--2383, 2021.

\bibitem[Onozuka et~al.(2021)Onozuka, Matsumi, and
  Shino]{onozuka2021autonomous}
Yuya Onozuka, Ryosuke Matsumi, and Motoki Shino.
\newblock Autonomous mobile robot navigation independent of road boundary using
  driving recommendation map.
\newblock In \emph{2021 IEEE/RSJ International Conference on Intelligent Robots
  and Systems (IROS)}, pages 4501--4508. IEEE, 2021.

\bibitem[Oord et~al.(2018)Oord, Li, and Vinyals]{oord2018representation}
Aaron van~den Oord, Yazhe Li, and Oriol Vinyals.
\newblock Representation learning with contrastive predictive coding.
\newblock \emph{arXiv preprint arXiv:1807.03748}, 2018.

\bibitem[Pan et~al.(2020)Pan, Shin, Rameau, Lee, and
  Kweon]{pan2020unsupervised}
Fei Pan, Inkyu Shin, Francois Rameau, Seokju Lee, and In~So Kweon.
\newblock Unsupervised intra-domain adaptation for semantic segmentation
  through self-supervision.
\newblock In \emph{Proceedings of the IEEE/CVF Conference on Computer Vision
  and Pattern Recognition}, pages 3764--3773, 2020.

\bibitem[Pan et~al.(2018)Pan, Luo, Shi, and Tang]{pan2018two}
Xingang Pan, Ping Luo, Jianping Shi, and Xiaoou Tang.
\newblock Two at once: Enhancing learning and generalization capacities via
  ibn-net.
\newblock In \emph{Proceedings of the European Conference on Computer Vision
  (ECCV)}, pages 464--479, 2018.

\bibitem[Pan et~al.(2019)Pan, Zhan, Shi, Tang, and Luo]{pan2019switchable}
Xingang Pan, Xiaohang Zhan, Jianping Shi, Xiaoou Tang, and Ping Luo.
\newblock Switchable whitening for deep representation learning.
\newblock In \emph{Proceedings of the IEEE/CVF International Conference on
  Computer Vision}, pages 1863--1871, 2019.

\bibitem[Park et~al.(2020)Park, Efros, Zhang, and Zhu]{park2020contrastive}
Taesung Park, Alexei~A Efros, Richard Zhang, and Jun-Yan Zhu.
\newblock Contrastive learning for unpaired image-to-image translation.
\newblock In \emph{Computer Vision--ECCV 2020: 16th European Conference,
  Glasgow, UK, August 23--28, 2020, Proceedings, Part IX 16}, pages 319--345.
  Springer, 2020.

\bibitem[Peng et~al.(2021)Peng, Lei, Liu, Zhang, and Liu]{peng2021global}
Duo Peng, Yinjie Lei, Lingqiao Liu, Pingping Zhang, and Jun Liu.
\newblock Global and local texture randomization for synthetic-to-real semantic
  segmentation.
\newblock \emph{IEEE Transactions on Image Processing}, 30:\penalty0
  6594--6608, 2021.

\bibitem[Peng et~al.(2022)Peng, Lei, Hayat, Guo, and Li]{peng2022semantic}
Duo Peng, Yinjie Lei, Munawar Hayat, Yulan Guo, and Wen Li.
\newblock Semantic-aware domain generalized segmentation.
\newblock In \emph{Proceedings of the IEEE/CVF Conference on Computer Vision
  and Pattern Recognition}, pages 2594--2605, 2022.

\bibitem[Richter et~al.(2016)Richter, Vineet, Roth, and
  Koltun]{richter2016playing}
Stephan~R Richter, Vibhav Vineet, Stefan Roth, and Vladlen Koltun.
\newblock Playing for data: Ground truth from computer games.
\newblock In \emph{Computer Vision--ECCV 2016: 14th European Conference,
  Amsterdam, The Netherlands, October 11-14, 2016, Proceedings, Part II 14},
  pages 102--118. Springer, 2016.

\bibitem[Ronneberger et~al.(2015)Ronneberger, Fischer, and
  Brox]{ronneberger2015u}
Olaf Ronneberger, Philipp Fischer, and Thomas Brox.
\newblock U-net: Convolutional networks for biomedical image segmentation.
\newblock In \emph{Medical Image Computing and Computer-Assisted
  Intervention--MICCAI 2015: 18th International Conference, Munich, Germany,
  October 5-9, 2015, Proceedings, Part III 18}, pages 234--241. Springer, 2015.

\bibitem[Ros et~al.(2016)Ros, Sellart, Materzynska, Vazquez, and
  Lopez]{ros2016synthia}
German Ros, Laura Sellart, Joanna Materzynska, David Vazquez, and Antonio~M
  Lopez.
\newblock The synthia dataset: A large collection of synthetic images for
  semantic segmentation of urban scenes.
\newblock In \emph{Proceedings of the IEEE conference on computer vision and
  pattern recognition}, pages 3234--3243, 2016.

\bibitem[Sakaridis et~al.(2021)Sakaridis, Dai, and Van~Gool]{sakaridis2021acdc}
Christos Sakaridis, Dengxin Dai, and Luc Van~Gool.
\newblock Acdc: The adverse conditions dataset with correspondences for
  semantic driving scene understanding.
\newblock In \emph{Proceedings of the IEEE/CVF International Conference on
  Computer Vision}, pages 10765--10775, 2021.

\bibitem[Sandler et~al.(2018)Sandler, Howard, Zhu, Zhmoginov, and
  Chen]{sandler2018mobilenetv2}
Mark Sandler, Andrew Howard, Menglong Zhu, Andrey Zhmoginov, and Liang-Chieh
  Chen.
\newblock Mobilenetv2: Inverted residuals and linear bottlenecks.
\newblock In \emph{Proceedings of the IEEE conference on computer vision and
  pattern recognition}, pages 4510--4520, 2018.

\bibitem[Ulyanov et~al.(2016)Ulyanov, Vedaldi, and
  Lempitsky]{ulyanov2016instance}
Dmitry Ulyanov, Andrea Vedaldi, and Victor Lempitsky.
\newblock Instance normalization: The missing ingredient for fast stylization.
\newblock \emph{arXiv preprint arXiv:1607.08022}, 2016.

\bibitem[Van~der Maaten and Hinton(2008)]{van2008visualizing}
Laurens Van~der Maaten and Geoffrey Hinton.
\newblock Visualizing data using t-sne.
\newblock \emph{Journal of machine learning research}, 9\penalty0 (11), 2008.

\bibitem[Volpi et~al.(2018)Volpi, Namkoong, Sener, Duchi, Murino, and
  Savarese]{volpi2018generalizing}
Riccardo Volpi, Hongseok Namkoong, Ozan Sener, John~C Duchi, Vittorio Murino,
  and Silvio Savarese.
\newblock Generalizing to unseen domains via adversarial data augmentation.
\newblock \emph{Advances in neural information processing systems}, 31, 2018.

\bibitem[Vu et~al.(2019)Vu, Jain, Bucher, Cord, and P{\'e}rez]{vu2019advent}
Tuan-Hung Vu, Himalaya Jain, Maxime Bucher, Matthieu Cord, and Patrick
  P{\'e}rez.
\newblock Advent: Adversarial entropy minimization for domain adaptation in
  semantic segmentation.
\newblock In \emph{Proceedings of the IEEE/CVF conference on computer vision
  and pattern recognition}, pages 2517--2526, 2019.

\bibitem[Wang et~al.(2021)Wang, Zhou, Yu, Dai, Konukoglu, and
  Van~Gool]{wang2021exploring}
Wenguan Wang, Tianfei Zhou, Fisher Yu, Jifeng Dai, Ender Konukoglu, and Luc
  Van~Gool.
\newblock Exploring cross-image pixel contrast for semantic segmentation.
\newblock In \emph{Proceedings of the IEEE/CVF International Conference on
  Computer Vision}, pages 7303--7313, 2021.

\bibitem[Wu et~al.(2022)Wu, Wu, Zhang, Ju, and Wang]{wu2022siamdoge}
Zhenyao Wu, Xinyi Wu, Xiaoping Zhang, Lili Ju, and Song Wang.
\newblock Siamdoge: Domain generalizable semantic segmentation using siamese
  network.
\newblock In \emph{European Conference on Computer Vision}, pages 603--620.
  Springer, 2022.

\bibitem[Xie et~al.(2021)Xie, Wang, Yu, Anandkumar, Alvarez, and
  Luo]{xie2021segformer}
Enze Xie, Wenhai Wang, Zhiding Yu, Anima Anandkumar, Jose~M Alvarez, and Ping
  Luo.
\newblock Segformer: Simple and efficient design for semantic segmentation with
  transformers.
\newblock \emph{Advances in Neural Information Processing Systems},
  34:\penalty0 12077--12090, 2021.

\bibitem[Xu et~al.(2022)Xu, Yao, Jiang, Jiang, Chu, Han, Zhang, Wang, and
  Tai]{xu2022dirl}
Qi Xu, Liang Yao, Zhengkai Jiang, Guannan Jiang, Wenqing Chu, Wenhui Han, Wei
  Zhang, Chengjie Wang, and Ying Tai.
\newblock Dirl: Domain-invariant representation learning for generalizable
  semantic segmentation.
\newblock In \emph{Proceedings of the AAAI Conference on Artificial
  Intelligence}, pages 2884--2892, 2022.

\bibitem[Yang et~al.(2023)Yang, Gu, and Sun]{yang2023generalized}
Liwei Yang, Xiang Gu, and Jian Sun.
\newblock Generalized semantic segmentation by self-supervised source domain
  projection and multi-level contrastive learning.
\newblock \emph{arXiv preprint arXiv:2303.01906}, 2023.

\bibitem[Yoo et~al.(2019)Yoo, Uh, Chun, Kang, and Ha]{yoo2019photorealistic}
Jaejun Yoo, Youngjung Uh, Sanghyuk Chun, Byeongkyu Kang, and Jung-Woo Ha.
\newblock Photorealistic style transfer via wavelet transforms.
\newblock In \emph{Proceedings of the IEEE/CVF International Conference on
  Computer Vision}, pages 9036--9045, 2019.

\bibitem[Yu et~al.(2020)Yu, Chen, Wang, Xian, Chen, Liu, Madhavan, and
  Darrell]{yu2020bdd100k}
Fisher Yu, Haofeng Chen, Xin Wang, Wenqi Xian, Yingying Chen, Fangchen Liu,
  Vashisht Madhavan, and Trevor Darrell.
\newblock Bdd100k: A diverse driving dataset for heterogeneous multitask
  learning.
\newblock In \emph{Proceedings of the IEEE/CVF conference on computer vision
  and pattern recognition}, pages 2636--2645, 2020.

\bibitem[Yu et~al.(2021)Yu, Zhang, Dong, Hu, Dong, and Zhang]{yu2021dast}
Fei Yu, Mo Zhang, Hexin Dong, Sheng Hu, Bin Dong, and Li Zhang.
\newblock Dast: Unsupervised domain adaptation in semantic segmentation based
  on discriminator attention and self-training.
\newblock In \emph{Proceedings of the AAAI Conference on Artificial
  Intelligence}, pages 10754--10762, 2021.

\bibitem[Yue et~al.(2019)Yue, Zhang, Zhao, Sangiovanni-Vincentelli, Keutzer,
  and Gong]{yue2019domain}
Xiangyu Yue, Yang Zhang, Sicheng Zhao, Alberto Sangiovanni-Vincentelli, Kurt
  Keutzer, and Boqing Gong.
\newblock Domain randomization and pyramid consistency: Simulation-to-real
  generalization without accessing target domain data.
\newblock In \emph{Proceedings of the IEEE/CVF International Conference on
  Computer Vision}, pages 2100--2110, 2019.

\bibitem[Zhou et~al.(2020)Zhou, Yang, Qiao, and Xiang]{zhou2020domain}
Kaiyang Zhou, Yongxin Yang, Yu Qiao, and Tao Xiang.
\newblock Domain generalization with mixstyle.
\newblock In \emph{International Conference on Learning Representations}, 2020.

\bibitem[Zou et~al.(2018)Zou, Yu, Kumar, and Wang]{zou2018unsupervised}
Yang Zou, Zhiding Yu, BVK Kumar, and Jinsong Wang.
\newblock Unsupervised domain adaptation for semantic segmentation via
  class-balanced self-training.
\newblock In \emph{Proceedings of the European conference on computer vision
  (ECCV)}, pages 289--305, 2018.

\end{thebibliography}
}

\appendix
\clearpage
\setcounter{page}{1}
\maketitlesupplementary

\section{Implementation Details of BlindNet}
\begin{figure}
  \centering
  \includegraphics[width=0.8\linewidth]{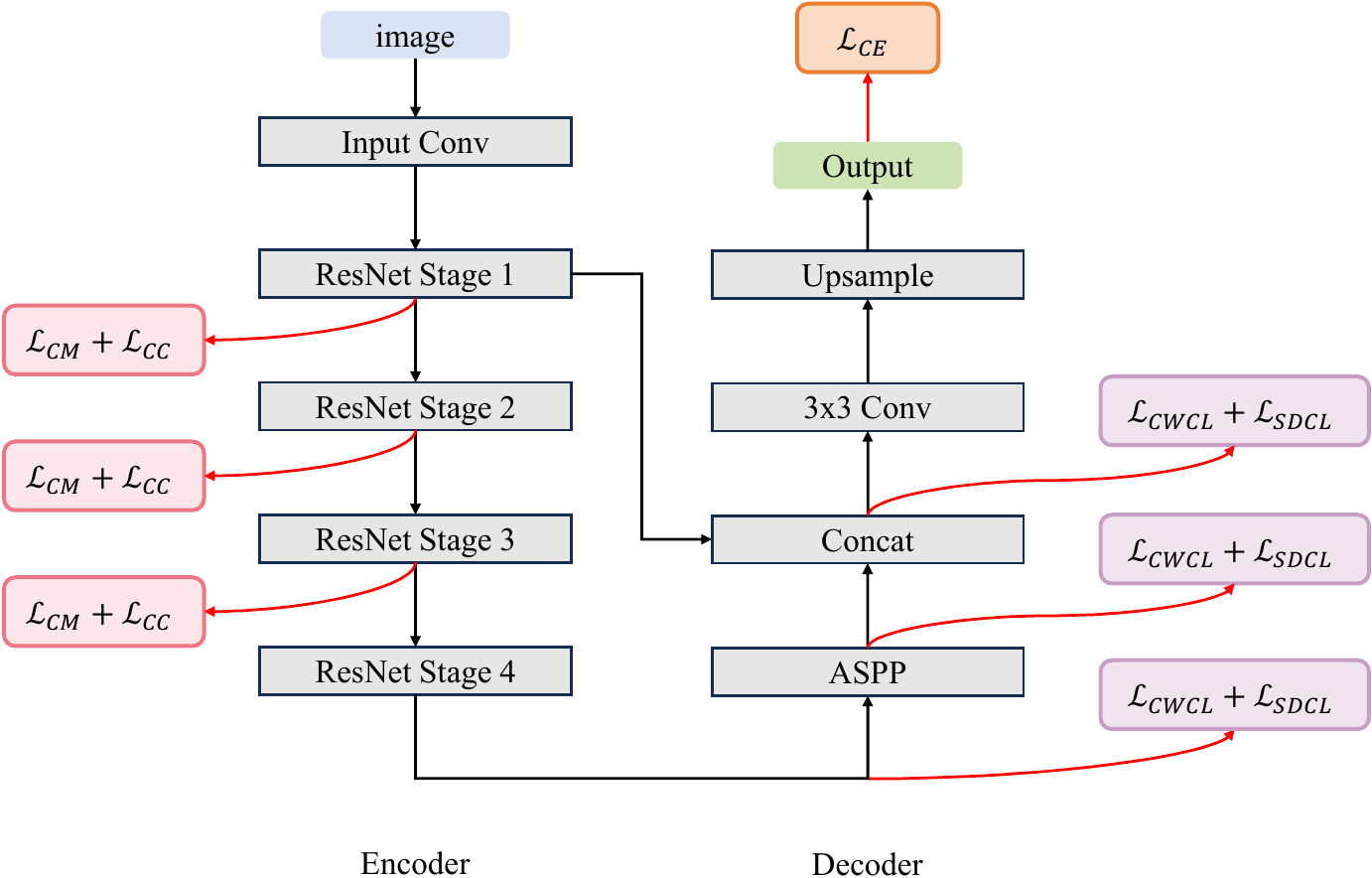}
  \caption{Implementation details of proposed loss functions in the DeepLabV3+ architecture.}
  \label{fig:arch}
\end{figure}

As shown in Fig.~\ref{fig:arch}, we apply our covariance alignment losses to the encoder features and the semantic consistency contrastive learning to the decoder features.

\section{More Results}
In this section, we show the detailed quantitative compassion results (Section~\ref{sec:app_qunat}) and additional qualitative results (Section~\ref{sec:app_qualit}) of our study.

\subsection{Quantitative Results}
\label{sec:app_qunat}

Table~\ref{tab:class} reports a comparison of pixel accuracy and IoU for each semantic class between DGSS methods. Our model significantly outperforms others in overall pixel accuracy, indicating its robust performance. In IoU for each semantic class, our model particularly excels in roads, sidewalks, sky, people, riders, and cars, which are commonly present in photos.  However, the table also indicates a degraded performance in classes such as traffic signs, traffic lights, and trains, which are less frequently encountered in the source domain (GTAV). Our future work will aim to address this issue and improve performance across all classes.

\subsection{Qualitative Results}
\label{sec:app_qualit}

\begin{figure*}
  \centering
  \includegraphics[width=\linewidth]{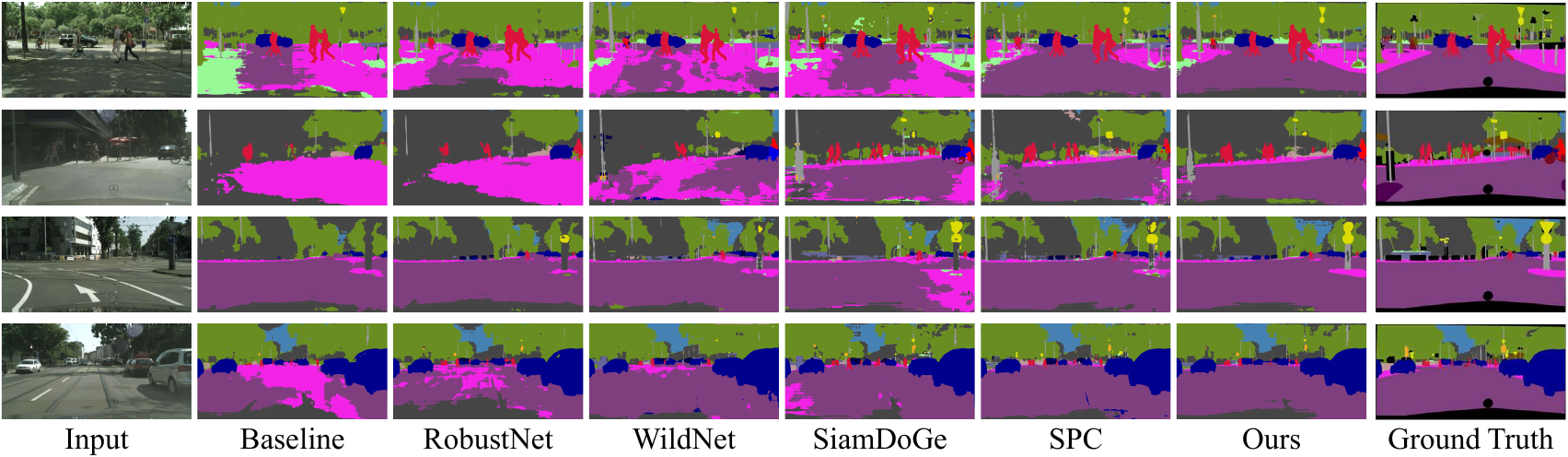}
  \caption{Qualitative comparison between DGSS methods trained on GTAV (G) and tested on unseen target domains of Cityscapes (C) using DeeplabV3+ with ResNet50 backbone.}
  \label{fig:city}
\end{figure*}

\begin{figure*}
  \centering
  \includegraphics[width=\linewidth]{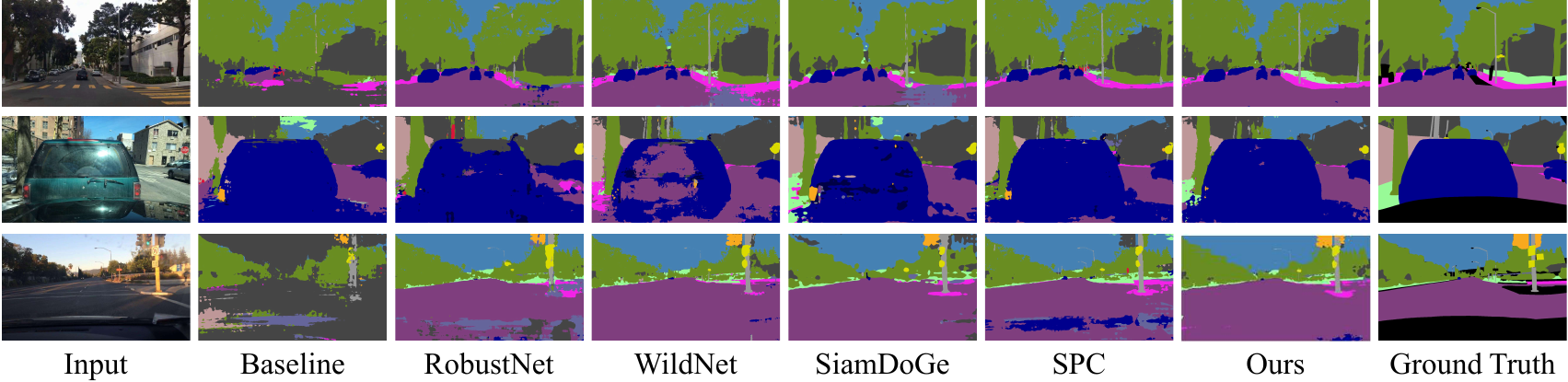}
  \caption{Qualitative comparison between DGSS methods trained on GTAV (G) and tested on unseen target domains of BDD100K (B) using DeeplabV3+ with ResNet50 backbone.}
  \label{fig:bdd}
\end{figure*}

\begin{figure*}
  \centering
  \includegraphics[width=\linewidth]{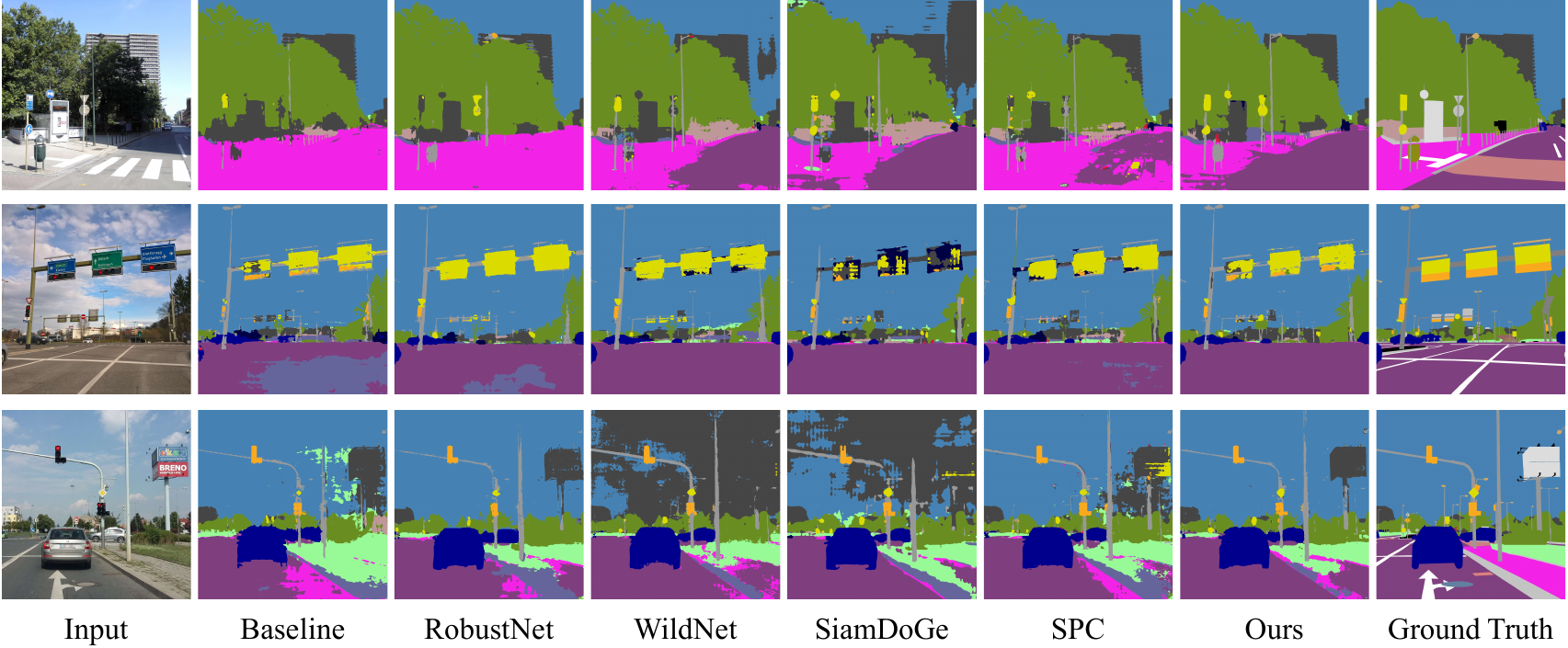}
  \caption{Qualitative comparison between DGSS methods trained on GTAV (G) and tested on unseen target domains of Mapillary (M) using DeeplabV3+ with ResNet50 backbone.}
  \label{fig:map}
\end{figure*}

Figs.~\ref{fig:city} (G$\xrightarrow{}$C), \ref{fig:bdd} (G$\xrightarrow{}$B), and \ref{fig:map} (G$\xrightarrow{}$M) present qualitative comparisons between our model and others, including baseline~\cite{chen2017rethinking}, RobustNet~\cite{choi2021robustnet}, WildNet~\cite{lee2022wildnet}, SiamDoGe~\cite{wu2022siamdoge}, and SPC~\cite{huang2023style}. The results clearly illustrate our model's consistent superiority, particularly in the segmentation of sidewalks, roads, buildings, terrain, and cars. The result demonstrates the robustness and effectiveness of our model in handling DGSS.

\section{More Ablation Studies}
In this section, we conduct more ablation studies on our model. In Section~\ref{sec:app_ab_qualit}, we show a qualitative analysis of the proposed loss functions, and in Section~\ref{sec:app_ab_hpyerp}, we experiment on the weight of the proposed loss functions.

\subsection{Qualitative Results}
\label{sec:app_ab_qualit}

\begin{figure*}
  \centering
  \includegraphics[width=\linewidth]{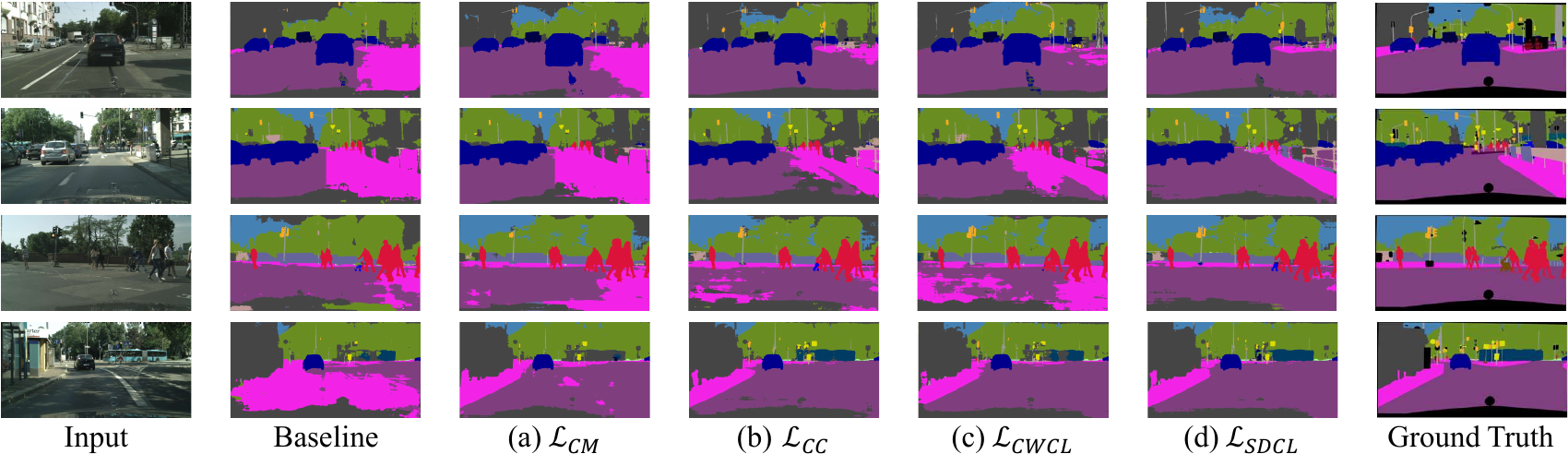}
  \caption{Qualitative comparison for ablation studies. The models are trained on GTAV (G) and tested on unseen target domains of Mapillary (M) using DeeplabV3+ with ResNet50 backbone. (a) $\mathcal{L}_{CM}$. (b) $\mathcal{L}_{CM}+\mathcal{L}_{CC}$. (c) $\mathcal{L}_{CM}+\mathcal{L}_{CC}+\mathcal{L}_{CWCL}$, (d) $\mathcal{L}_{CM}+\mathcal{L}_{CC}+\mathcal{L}_{CWCL}+\mathcal{L}_{SDCL}$}
  \label{fig:ablation}
\end{figure*}

We incrementally added each loss function ($\mathcal{L}_{CM}$, $\mathcal{L}_{CC}$, $\mathcal{L}_{CWCL}$, $\mathcal{L}_{SDCL}$) to the baseline model to validate the impact of loss. Fig.~\ref{fig:ablation} presents the qualitative results of the ablation studies on the proposed loss functions.

For our qualitative ablation study, we added each loss function ($\mathcal{L}_{CM}$, $\mathcal{L}_{CC}$, $\mathcal{L}_{CWCL}$, $\mathcal{L}_{SDCL}$) to the baseline model, validating their contributions. The results are depicted in Fig.~\ref{fig:ablation}. The introduction of CML ($\mathcal{L}_{CM}$) enhances the capture of the details such as traffic lights, as illustrated in Fig.~\ref{fig:ablation} row 1. Adding  CCL ($\mathcal{L}_{CC}$) further strengthens content representation, leading to an improvement in overall accuracy. The CWCL ($\mathcal{L}_{CWCL}$) strengthens semantic understanding, allowing for better detection of smaller objects. However, this enhancement comes with a trade-off, as it introduces some degree of confusion among similar classes (\eg sidewalk and road). The application of SDCL ($\mathcal{L}_{SDCL}$) effectively disentangles misclassified features, leading to clearer class distinctions.

\subsection{Hyper-parameter}
\label{sec:app_ab_hpyerp}

We varied the weighting parameters for each loss function in (\ref{eq:total}), and conducted experiments by adjusting each loss weight by 0.1, using the model configuration that initially showed the best performance as our baseline, reported in Table~\ref{tab:weighting}. The CML ($\mathcal{L}_{CM}$), a key component for style blindness, shows that an overly strong influence can significantly degrade network performance. Conversely, the CCL ($\mathcal{L}_{CC}$) and the CWCL ($\mathcal{L}_{CWCL}$) exhibit improved performance with a slightly higher influence than a lower influence.

\addtocounter{table}{1}
\begin{table}[]
\resizebox{\linewidth}{!}{%
\begin{tabular}{cc|cc|cccc}
\hline
$\omega_1$ ($\mathcal{L}_{CM}$) & $\omega_2$ ($\mathcal{L}_{CC}$) & $\omega_3$ ($\mathcal{L}_{CWCL}$) & $\omega_4$ ($\mathcal{L}_{SDCL}$) & C & B & M & S \\ \hline
0.2 & 0.2 & 0.3 & 0.3 & \textbf{45.72} & \textbf{41.32} & \underline{47.08} & \textbf{31.39}\\ \hline
\textbf{0.1} & 0.2 & 0.3 & 0.3 & 45.06 & 39.37 & 45.14 & \underline{31.09}\\
\textbf{0.3} & 0.2 & 0.3 & 0.3 & 43.04 & 38.75 & 44.69 & 29.58\\ \hline
0.2 & \textbf{0.1} & 0.3 & 0.3 & 44.15 & 39.15 & 46.00 & 30.62 \\
0.2 & \textbf{0.3} & 0.3 & 0.3 & 44.78 & 40.01 & 46.56 & 30.74\\ \hline
0.2 & 0.2 & \textbf{0.2} & 0.3 & 43.42 & 39.24 & 45.55 & 30.40\\
0.2 & 0.2 & \textbf{0.4} & 0.3 & \underline{45.52} & 39.88 & 45.73 & 30.20\\ \hline
0.2 & 0.2 & 0.3 & \textbf{0.2} & 44.58 & \underline{40.42} & \textbf{47.35} & 30.72 \\
0.2 & 0.2 & 0.3 & \textbf{0.4} & 45.26 & 40.16 & 46.91 & 30.49\\ \hline
\end{tabular}%
}
\caption{Sensitivity to weighting parameters of each loss function}
\label{tab:weighting}
\end{table}

\newpage

\addtocounter{table}{-2}
\begin{table*}[h]
\resizebox{\textwidth}{!}{%
\begin{tabular}{l|cc|ccccccccccccccccccc}
\hline
 &&&
\cellcolor[RGB]{128,64,128} & 
\cellcolor[RGB]{244,35,232} & 
\cellcolor[RGB]{70,70,70} & 
\cellcolor[RGB]{102,102,156} & 
\cellcolor[RGB]{190,153,153} & 
\cellcolor[RGB]{153,153,153} & 
\cellcolor[RGB]{250,170,30} & 
\cellcolor[RGB]{220,220,0} & 
\cellcolor[RGB]{107,142,35} & 
\cellcolor[RGB]{152,251,152} & 
\cellcolor[RGB]{70,130,180} & 
\cellcolor[RGB]{220,20,60} & 
\cellcolor[RGB]{255,0,0} & 
\cellcolor[RGB]{0,0,142} & 
\cellcolor[RGB]{0,0,70} & 
\cellcolor[RGB]{0,60,100} & 
\cellcolor[RGB]{0,80,100} & 
\cellcolor[RGB]{0,0,230} & 
\cellcolor[RGB]{119,11,32} \\
Methods &
\begin{tabular}[c]{@{}c@{}}Pixel\\ Accuracy\end{tabular}&
  mIoU &
 \rotatebox[origin=l]{90}{Road} &
  \rotatebox[origin=l]{90}{Sidewalk} &
  \rotatebox[origin=l]{90}{Building} &
  \rotatebox[origin=l]{90}{Wall} &
  \rotatebox[origin=l]{90}{Fence} &
  \rotatebox[origin=l]{90}{Pole} &
  \rotatebox[origin=l]{90}{Traffic light \quad} &
  \rotatebox[origin=l]{90}{Traffic sign} &
  \rotatebox[origin=l]{90}{Vegetation} &
  \rotatebox[origin=l]{90}{Terrain} &
  \rotatebox[origin=l]{90}{Sky} &
  \rotatebox[origin=l]{90}{Person} &
  \rotatebox[origin=l]{90}{Rider} &
  \rotatebox[origin=l]{90}{Car} &
  \rotatebox[origin=l]{90}{Truck} &
  \rotatebox[origin=l]{90}{Bus} &
  \rotatebox[origin=l]{90}{Train} &
  \rotatebox[origin=l]{90}{Motorcycle} &
  \rotatebox[origin=l]{90}{Bicycle} \\ \hline \hline
Baseline~\cite{chen2017rethinking} & 71.02 &  29.0 &  51.9 &  20.6 &  57.2 &  22.4 &  21.0 &  25.3 &  24.9 &  10.1 &  61.3 &  23.7 &  52.0 &
  53.8 &  13.6 &  51.2 &  19.5 &  21.2 &  0.3 &  12.0 &  8.1  \\
RobustNet~\cite{choi2021robustnet}& 77.18 &  37.3 &  58.9 &  27.7 &  63.2 &  22.8 &  23.1 &  26.4 &  30.6 &  20.7 &  85.1 &  39.2 &  69.8 &  62.4 &  15.9 &  76.7 &  23.2 &  22.3 &  3.9 &  18.4 &  18.6  \\
SiamDoGe~\cite{wu2022siamdoge}& 84.73 &  43.0  &  83.7 &  34.1 &  78.6 &  26.4 &  25.6 &  26.0 &  \textbf{42.4} &  \textbf{28.6} &  84.3 &  28.1 &  68.9 &  62.1 &  \underline{31.1} &  \underline{85.6} &  31.3 &  28.9 &  3.5 &  22.8 &  23.3 \\
WildNet~\cite{lee2022wildnet}& 84.57 &  44.6 &  81.2 &  \underline{38.2} &  76.9 &  28.1 &  25.1 &  35.1 &  32.1 &  24.5 &  \textbf{85.4} &  35.4 &  72.2 &  \underline{65.0} &  27.3 &  85.5 &  29.7 &  33.2 &  \textbf{12.6} &  \textbf{32.8} &  \underline{27.4} \\
SPC~\cite{huang2023style} & \underline{86.65} &  44.1 &  \underline{86.9} &  37.8 &  \underline{81.2} &  \underline{28.9} &  26.9 &  \textbf{36.9} &  35.1 &  25.2 &  83.7 &  36.2 &  {78.5} &  63.9 &  30.4 &  84.1 &  24.8 &  28.1 &  \underline{12.1} &  19.3 &  17.9\\
DPCL~\cite{yang2023generalized} & 82.22 &  \underline{44.7} &  75.6 &  32.8 &  73.2 &  26.1 &  23.5 &  34.1 &  \underline{42.3} &  \underline{28.2} &  \underline{85.2} &  \underline{38.5} &   \textbf{81.2} &  {63.8} &  25.0 &  76.6 &  \underline{31.7} &  \underline{33.9} &  5.7 &  \underline{27.6} &  \underline{45.0}  \\
Ours& \textbf{87.91} &  \textbf{45.7} &  \textbf{88.3} &  \textbf{44.1} &  \textbf{82.4} &  \textbf{30.9} &  \textbf{26.8} &  \underline{35.4} &  {33.4} &  {20.3} &  {85.0} &  \textbf{34.2} &
  \underline{78.5} &  \textbf{66.0} &  \textbf{33.7} &  \textbf{86.8} &  \textbf{33.0} &  \textbf{41.1} &  {1.4} &  {25.3} &  {22.1}  \\ \hline
\end{tabular}%
}
\caption{Quantitative results for pixel accuracy and each semantic class. The models are trained on GTAV and tested on Cityscapes using a ResNet50 backbone. The best and second best results are \textbf{bolded} and \underline{underlined}, respectively}
\label{tab:class}
\end{table*}

\end{document}